\documentclass{article}

\usepackage[final]{timeseries_workshop}
\usepackage{natbib}
\usepackage{amsmath}
\usepackage{amssymb}
\usepackage{pifont}
\usepackage{microtype}
\usepackage{framed}
\usepackage{wrapfig}
\usepackage{mdframed}
\usepackage{csquotes}
\usepackage{hyperref}
\usepackage{tikz}
\usepackage{algorithm}
\usepackage{algorithmic}
\usepackage{subfig}
\usepackage{newfloat}
\usepackage{listings}
\usepackage{xspace} 
\usepackage{amsfonts}
\usepackage{amsthm}
\usepackage{microtype}
\usepackage{graphicx}
\usepackage{enumitem}
\usepackage{booktabs}
\usepackage{csquotes}
\usepackage{pifont}
\usepackage{makecell}
\usepackage{booktabs} 
\usepackage{algorithm}
\usepackage{graphicx}
\usepackage{subcaption} 
\usepackage{appendix} 
\usepackage{subfig}

\usepackage[utf8]{inputenc} 
\usepackage[T1]{fontenc}    
\usepackage{hyperref}       
\usepackage{url}      
\usepackage{csquotes}
\usepackage{booktabs}       
\usepackage{amsfonts}       
\usepackage{nicefrac}       
\usepackage{microtype}      
\usepackage{xcolor}         
\usepackage{graphicx}
\usepackage{soul} 

\definecolor{bgblue}{rgb}{0.85,0.92,1.0}   
\definecolor{bggreen}{rgb}{0.88,1.0,0.88}  
\definecolor{bgpink}{rgb}{1.0,0.85,0.93}   
\definecolor{bgpurple}{rgb}{0.80,0.85,1.0} 

\definecolor{textbrown}{rgb}{0.65,0.16,0.16} 
\definecolor{textblue}{rgb}{0.1,0.3,0.7}     
\definecolor{textpurple}{rgb}{0.5,0.2,0.5}   
\definecolor{textgreen}{rgb}{0.0,0.5,0.2}    

\renewcommand{\thefootnote}{\fnsymbol{footnote}}

\title{time2time: Causal Intervention in Hidden States to Simulate Rare Events in Time Series Foundation Models}

\author{Debdeep Sanyal\textsuperscript{1}, Aaryan Nagpal\textsuperscript{1}, Dhruv Kumar\textsuperscript{1, 2},\\ \textbf{Murari Mandal\textsuperscript{1, 3}, Saurabh Deshpande\textsuperscript{1,*}}  \\
\textsuperscript{1} Birla AI Labs, Office of Ananya Birla, \textsuperscript{2} BITS Pilani, \textsuperscript{3} KIIT Bhubaneswar
}

\begin{document}
\maketitle
\setcounter{footnote}{0}
\renewcommand{\thefootnote}{\fnsymbol{footnote}}
\footnotetext{Code available at \url{https://github.com/birla-ai-labs/time2time}}
\setcounter{footnote}{1}
\footnotetext{Correspondence: saurabh.deshpande-c@adityabirla.com}

\begin{abstract}


While transformer-based foundation models excel at forecasting routine patterns, two questions remain: do they internalize semantic concepts such as market regimes, or merely fit curves? And can their internal representations be leveraged to simulate rare, high-stakes events such as market crashes? To investigate this, we introduce activation transplantation, a causal intervention that manipulates hidden states by imposing the statistical moments of one event (e.g., a historical crash) onto another (e.g., a calm period) during the forward pass. This procedure deterministically steers forecasts: injecting crash semantics induces downturn predictions, while injecting calm semantics suppresses crashes and restores stability. Beyond binary control, we find that models encode a graded notion of event severity, with the latent vector norm directly correlating with the magnitude of systemic shocks. Validated across two architecturally distinct TSFMs, Toto (decoder only) and Chronos (encoder-decoder), our results demonstrate that steerable, semantically grounded representations are a robust property of large time series transformers. Our findings provide evidence for a \emph{latent concept space} that governs model predictions, shifting interpretability from post-hoc attribution to direct causal intervention, and enabling semantic \enquote{what-if} analysis for strategic stress-testing.

\end{abstract} 


\section{Introduction}

Transformer-based time series foundation models (TSFM) currently define the state-of-the-art in time series forecasting \cite{chronos,moirai,timesfm,toto,Yuqietal-2023-PatchTST}, yet their black-box nature poses critical risks and limits their utility in various analysis~\cite{bbrisk1,bbrisk2,bbrisk3}. In high stakes domains such as finance, healthcare, and energy, trustworthy forecasts are critical, as a single misinterpreted prediction can lead to catastrophic consequences. This raises a fundamental question: Do these models develop a genuine understanding of the processes they model, or are they merely sophisticated curve-fitters, parroting complex patterns without comprehending their meaning? Answering this is essential for building reliable systems in critical applications. 

Financial markets represent one of the most consequential domains, and thus serve as an ideal testbed for our study. This context leads us to a pivotal question: does a TSFM, pretrained on decades of market data, learn an internal, manipulable concept of a \enquote{market crash}? While prior work shows that TSFMs can represent low-level mathematical primitives like trends and seasonality and that they can be manipulated~\cite{related2,related1,related3}, it remains unknown if they can grasp holistic events defined not by simple functions, but by complex real-world dynamics like panic-driven volatility and cascading price declines.



To test for causation beyond mere correlation, we introduce a direct intervention. We probe our hypothesis that market regimes are encoded in population-level activation statistics (mean and std), and by transplanting these statistics from a periods with differing dynamics (calm or crash) mid-forward pass. This allows us to \enquote{implant} the signature of a crash, which, if our hypothesis holds, should deterministically steer the model's forecast to predict a change in it's forecast scale.

Our results provide a striking confirmation. Injecting the statistical signature of a major crash (e.g., 2000 \footnote{https://en.wikipedia.org/wiki/Dot-com\_bubble}, 2008 \footnote{https://en.wikipedia.org/wiki/2008\_financial\_crisis}, or 2020 \footnote{https://en.wikipedia.org/wiki/2020\_stock\_market\_crash}) forces a calm period's forecast into a sharp downturn. The converse intervention; imposing calm statistics onto a crash, suppresses the downturn forecast. Beyond this binary steering, we uncover a deeper nuance in the model's understanding: the model quantitatively encodes the relative severity of different crashes, as measured by the norm of its latent representation. The model has learned not just \textit{what} a crash is, but that the Dot-com crash of 2000 constitutes a more severe \textit{representational event} than the 2008 crisis, revealing a rich, semantically grounded internal world. Building on these results, this work makes the following contributions:

\begin{itemize}
\item \textbf{First direct, causal evidence of semantic concepts:} We provide the first causal evidence that TSFMs learn high-level semantic concepts of real-world events, moving beyond simple primitives to holistic concepts like market crashes.

\item \textbf{The discovery of nuanced, interpretable representations:} We discover that these learned concepts are remarkably nuanced, with event severity encoded as a continuous and interpretable latent variable (i.e., the activation vector norm).

\item \textbf{A path to controllable, risk-aware simulation:} Our findings establish a practical path toward a new class of controllable simulations, enabling practitioners to conduct risk-aware stress-testing by simulating the impact of historic systemic shocks.
\end{itemize}
\section{Causal Intervention via Activation Transplantation}
\begin{figure}[t]
\centering
    \includegraphics[width=0.8\textwidth]{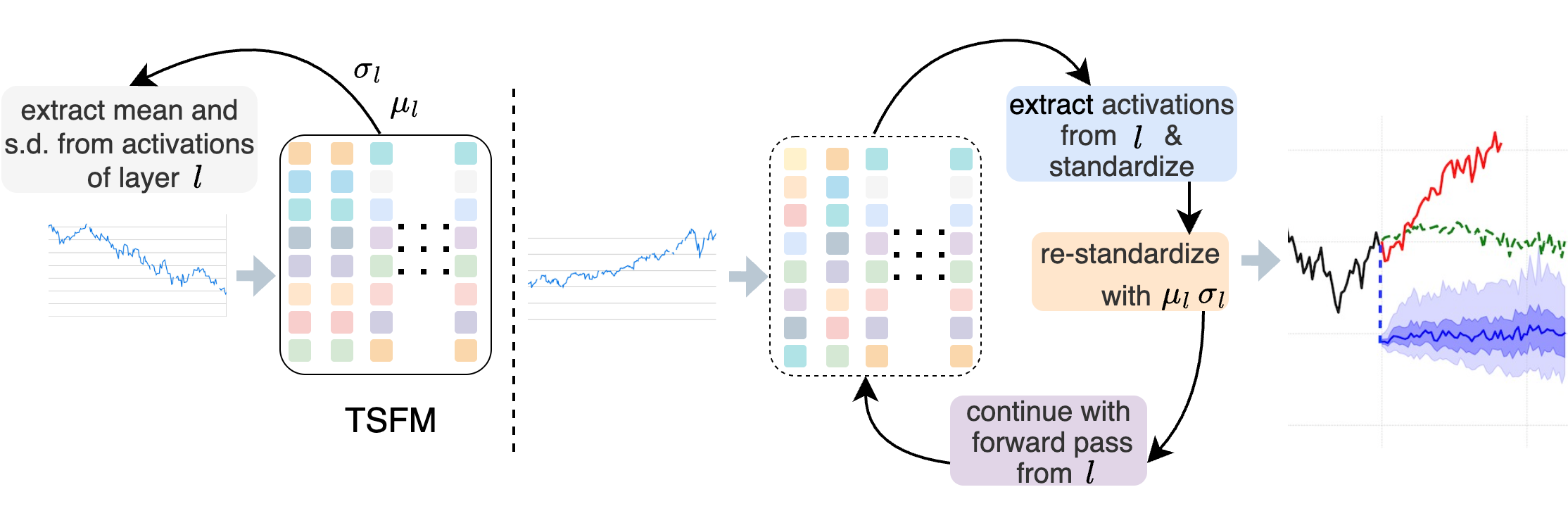}
    \caption{Overview of the proposed \textit{time2time} intervention. We extract the statistical moments (mean and standard deviation) of hidden activations at layer $\mathcal{l}$ from a style event, standardize the target activations, and re-standardize them with the style statistics. This activation transplantation implants the dynamics of one event (e.g., a market crash) into another (e.g., a calm period), steering the model’s forecast accordingly.}
    \label{fig:time2time}
    \vspace{-1em}
\end{figure}

\textbf{The Semantic Hypothesis in Activations} Let $M$ be a decoder-only TSFM with parameters $\theta$, consisting of $L$ layers. For a given input time series $X \in \mathbb{R}^{N \times T_{\text{in}}}$, where $N$ is the number of variates and $T_{\text{in}}$ is the length of the historical context, the model generates a forecast $\hat{Y} \in \mathbb{R}^{N \times T_{\text{out}}}$. The activation tensor at the output of any layer $l \in \{1, ..., L\}$ is denoted as:

\begin{equation}
A_l(X) = f_l(A_{l-1}(X); \theta_l) \in \mathbb{R}^{N \times T_{\text{in}} \times D}
\label{eq: forward pass}
\end{equation}

where $f_l$ is the function of the $l$-th Transformer block, $D$ is the hidden dimensionality, and $A_0(X)$ is the initial embedding of the input $X$. (For clarity, we omit the batch dimension from notation). Our guiding hypothesis is that the semantic signature of $X$ is encoded within the statistics of $A_l(X)$ for some layer $l$.

\textbf{The Transplantation Mechanism} Our procedure requires two distinct time series: a \textbf{style} series, $X_{\text{style}}$, and a \textbf{target} series, $X_{\text{target}}$. 

\textbf{Step 1: Extracting the Semantic Signature.} We first extract the semantic signature from the \textit{Style} data's activations at layer $l$. This signature is defined as the mean and standard deviation vectors computed \textbf{across the sequence length ($T_{\text{in}}$)}, capturing the characteristic neural statistics of the entire style period.

\begin{equation}
    \mu_l(X_{\text{style}}) = \frac{1}{T_{\text{in}}} \sum_{t=1}^{T_{\text{in}}} A_l(X_{\text{style}})_{:, t, :} \quad ; \quad \sigma_l(X_{\text{style}}) = \sqrt{\frac{1}{T_{\text{in}}} \sum_{t=1}^{T_{\text{in}}} (A_l(X_{\text{style}})_{:, t, :} - \mu_l)^2}
    \label{eq: style statistics}
\end{equation}

where the statistics are computed over the time-step index $t$, resulting in $\mu_l, \sigma_l \in \mathbb{R}^{N \times D}$. This operation directly corresponds to taking the mean and standard deviation along the sequence length dimension of the activation tensor.

\textbf{Step 2: Intervention on target Activations.} Next, in a separate forward pass with the \textit{target} data, we halt the computation after layer $l$. We then perform the core transplantation. We first standardize the target activations by their own statistics along sequence dimension to strip them of their original semantic signature. Then, we re-scale and shift them using the stored signature from the style data.

\begin{equation}    
\tilde{A}_l(X_{\text{target}}) = \left( \frac{A_l(X_{\text{target}}) - \mu_l(X_{\text{target}})}{\sigma_l(X_{\text{target}}) + \epsilon} \right) \odot \sigma_l(X_{\text{style}}) + \mu_l(X_{\text{style}})
\end{equation}

where $\odot$ denotes element-wise multiplication (with broadcasting) and $\epsilon$ is a small constant (e.g., $10^{-5}$) for numerical stability. This equation precisely replaces the time-averaged statistics of the target with those of the style, while preserving the time-step-varying structure of the normalized target.

\textbf{Step 3: Generating the Intervened Forecast.} Finally, we resume the forward pass from layer $l+1$, feeding the modified activation tensor $\tilde{A}_l$ into the rest of the network to produce a forecast conditioned on the implanted semantic concept.

\begin{equation}
\hat{Y}_{\text{intervened}} = M_{l+1 \rightarrow L}(\tilde{A}_l(X_{\text{target}}))
\end{equation}

\textit{This framework provides a general and powerful tool for causally testing and steering the conceptual representations within a TSFM model.}

\section{Experiments \& Results}
\label{sec:exp}

To rigorously test the generality of our claims, we perform our experiments on two TSFMs, deliberately chosen for their  architectural and scale diversity: Toto Open-Base-1.0 \cite{toto}, a 103M parameter decoder-only architecture, and four variants of Chronos, 8M-710M parameter encoder-decoder model.

\begin{figure}[t]
    \centering
    \subfloat[Toto-Open-Base-1.0
(decoder-only)]{%
        \includegraphics[width=0.515\textwidth]{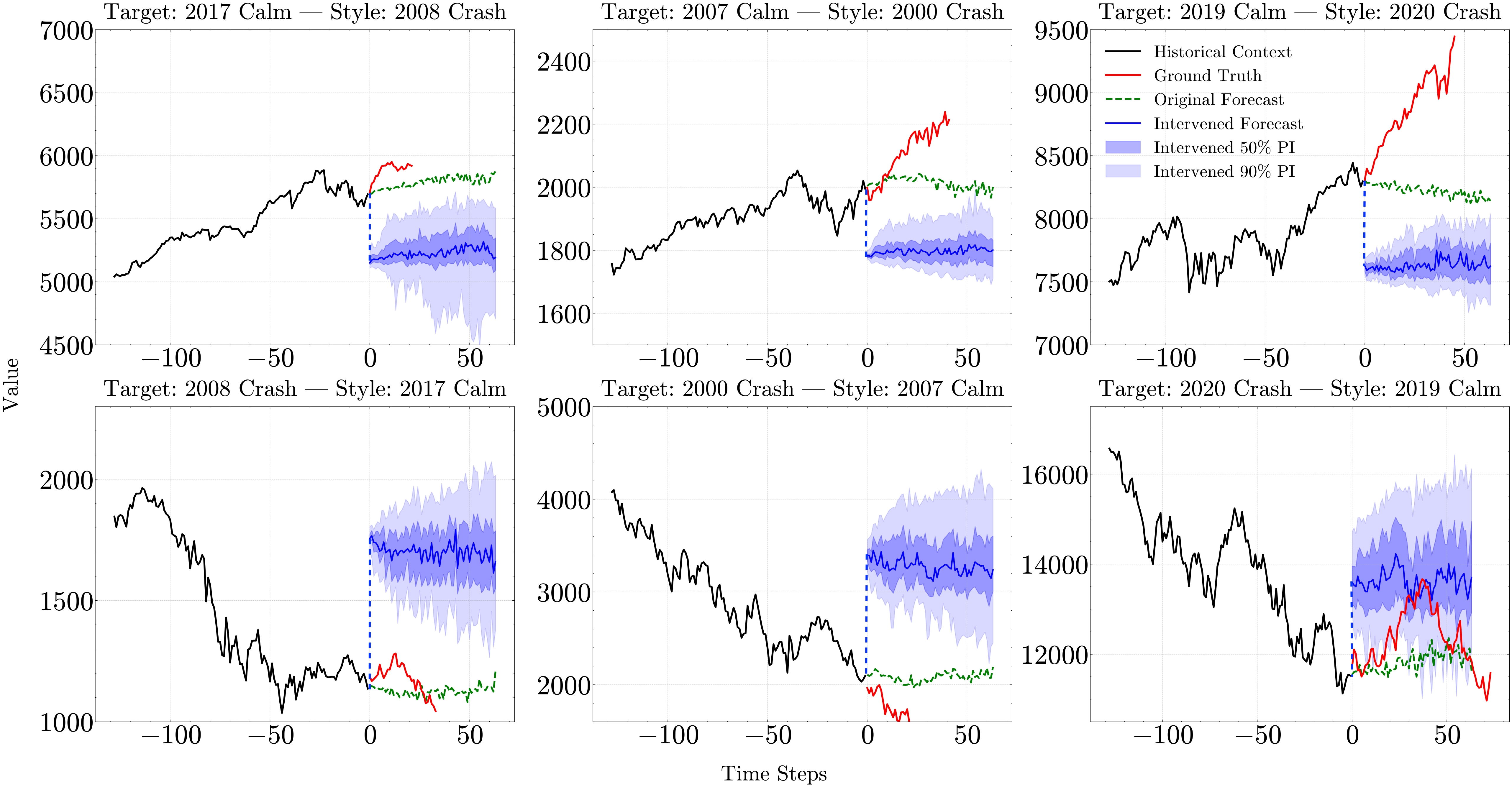}\label{fig: toto_intervention_crash_calm}
    }
    \hfill
    \subfloat[Chronos-small (encoder–decoder)]{%
        \includegraphics[width=0.46\textwidth]{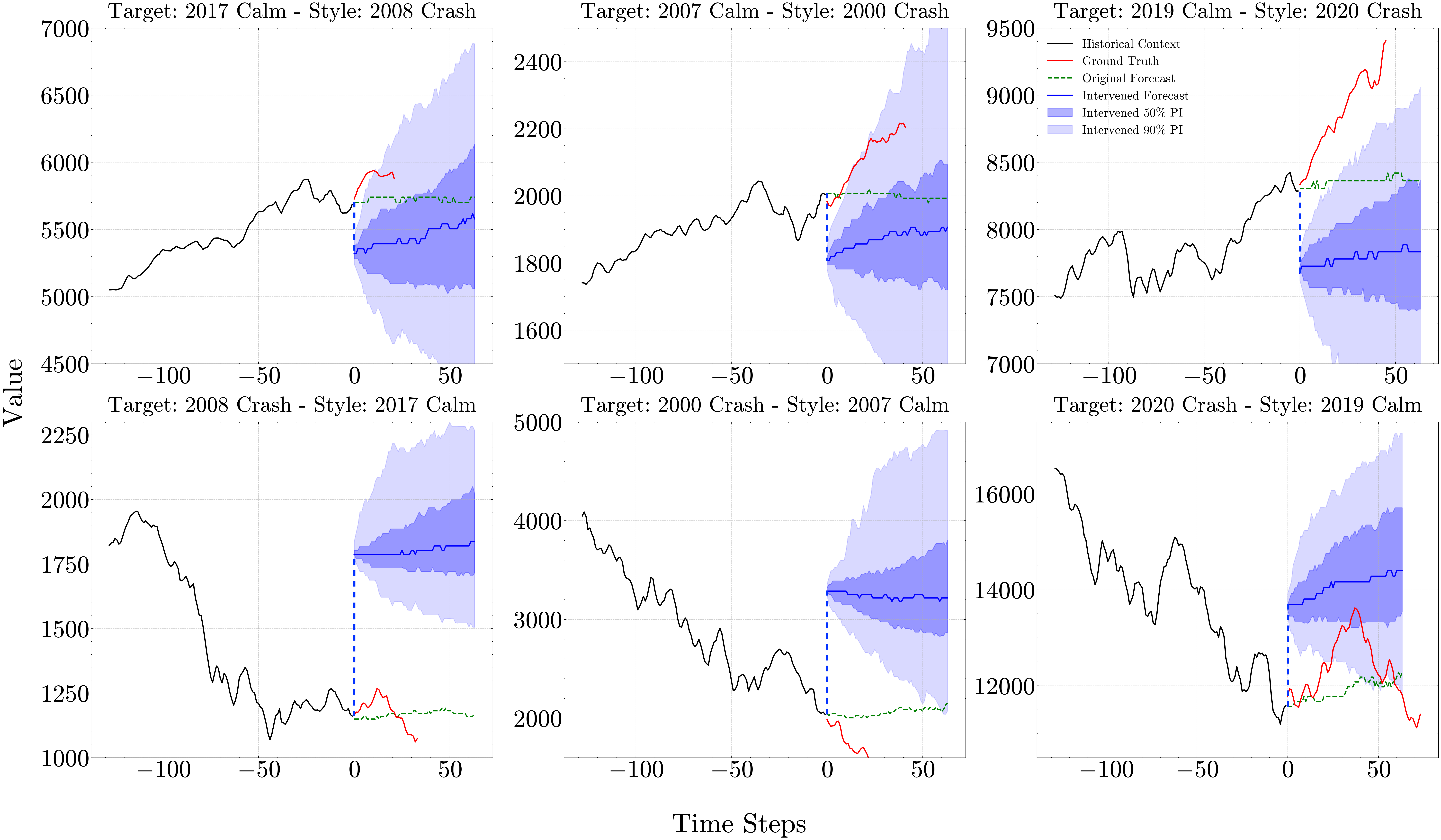}\label{fig: chronos_intervention_crash_calm}
    }
    \caption{\textbf{Forecast interventions via activation transplantation.} 
    We intervene on model forecasts at $l=8$ for both models by transferring statistical moments of hidden activations between regimes. \textbf{X-axis}: Time step in days \textbf{Y-axis}: NASDAQ 100 Index. \textbf{Top rows}: calm periods transplanted with crash statistics, which deterministically induce downturn forecasts simulating stress tests. 
    \textbf{Bottom rows}: crash periods transplanted with calm statistics, which suppress downturns and restore stability. 
    Shaded regions show 50\% and 90\% prediction intervals for the intervened forecasts, while green line indicates median forecasts by Toto and Chronos respectively. (Chronos Ablations in Section \ref{sec:chronos-ablations})}
    \label{fig:intervention_crash_calm}
    \vspace{-1em}
\end{figure}

\textbf{Activation transplantation provides direct, causal control over model forecasts.} 
Figure~\ref{fig:intervention_crash_calm} demonstrates that forecast reversals follow our interventions deterministically, moving beyond correlation to causal evidence. 
This control is bidirectional: imposing a \enquote{crash} signature on a \enquote{calm} context reliably induces sharp downturn forecasts, while injecting \enquote{calm} statistics into a \enquote{crash} offsets the forecast toward stability, counteracting the downturn implied by Toto and Chronos baselines (green). \textit{Notably, the offsets produced by both Toto and Chronos are of comparable magnitude for identical intervention cases.} Hence, these effects generalize across two architecturally distinct TSFMs, demonstrating that steerable crash and calm representations are not architecture-specific, but a robust property of large TSFMs. (Appendix \ref{sec:synth-ablation} for ablations with synthetic data).


\begin{figure*}[!ht]
    \centering
    \begin{minipage}[b]{0.4\textwidth}   
        \centering
        \subfloat[][Crash–Calm\label{fig:cal-crash-heatmap}]{
            \includegraphics[width=0.45\textwidth]{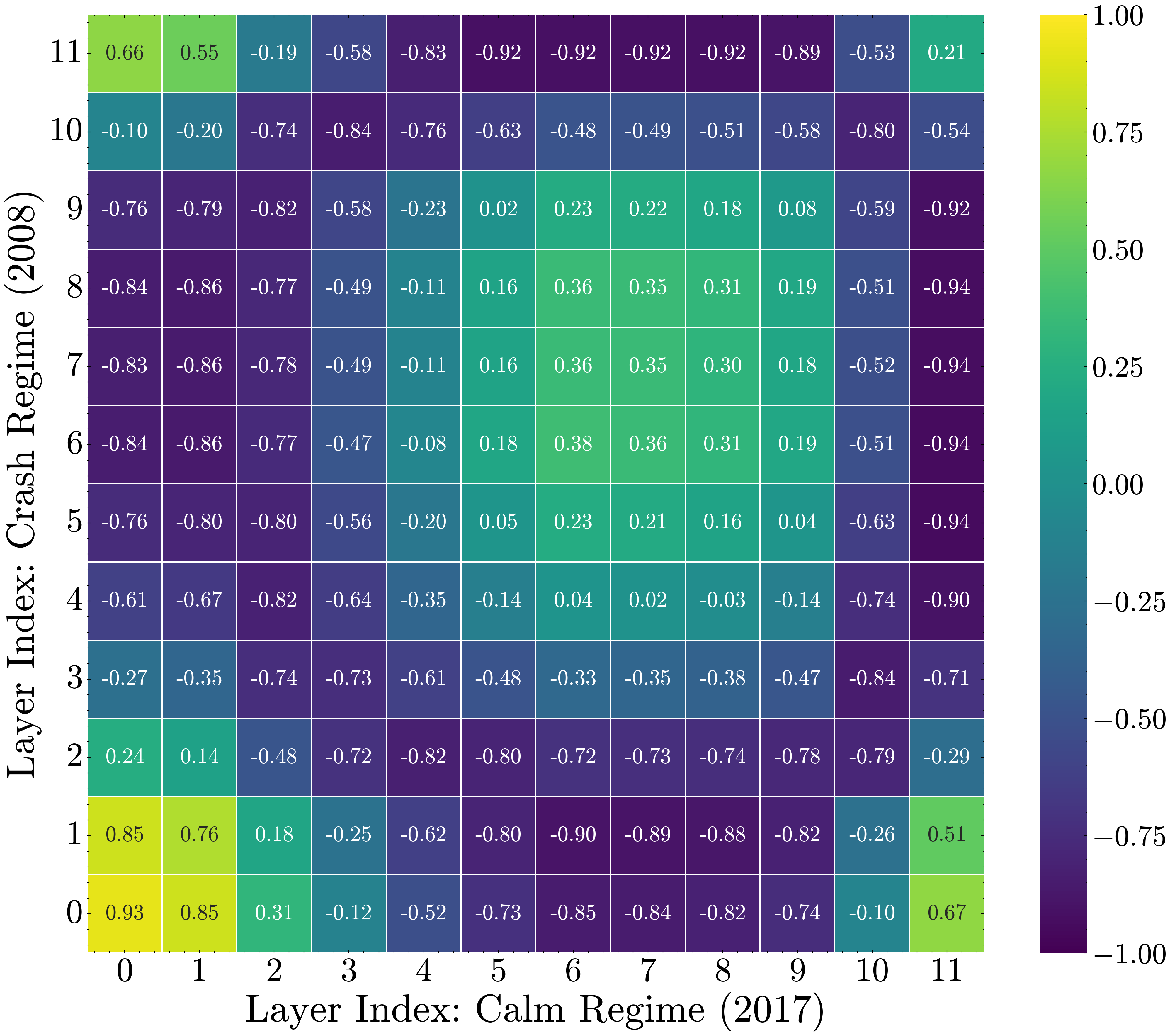}
        }
        \hspace{0.001\textwidth}
        \subfloat[][Crash–Crash\label{fig:crash-crash-heatmap}]{
            \includegraphics[width=0.45\textwidth]{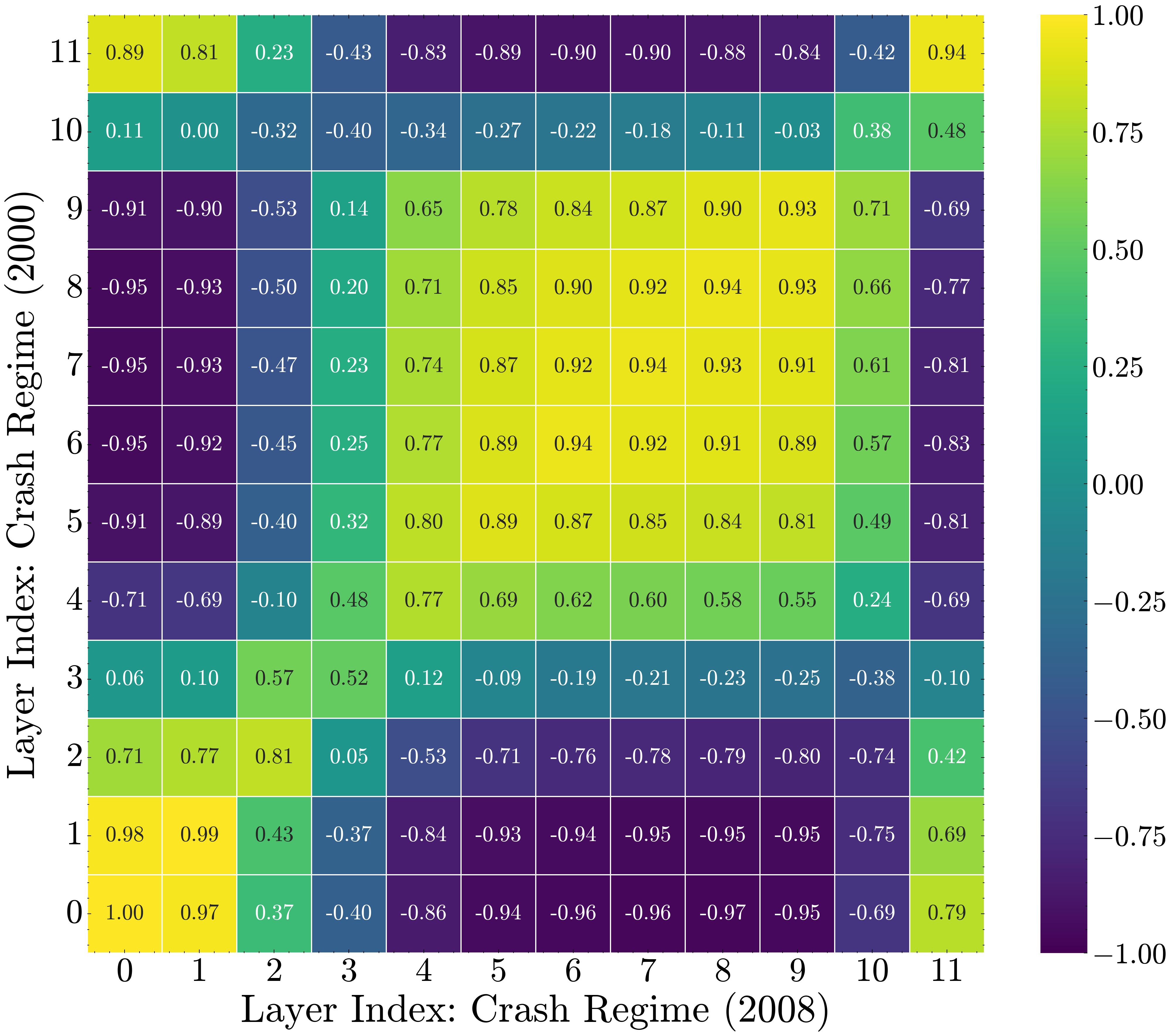}
        }
        \caption{Cross-regime similarity in reduced latent subspace. 
        (a) crash–calm pairs are strongly anti-correlated in early layers but gradually align. 
        (b) crash–crash pairs rapidly converge into a coherent latent subspace by mid layers.}
        \label{fig:heatmap-plots}
        \vspace{-0.5em}
    \end{minipage}%
    \hfill
    \begin{minipage}[b]{0.58\textwidth}   
        \centering
        \includegraphics[width=\textwidth]{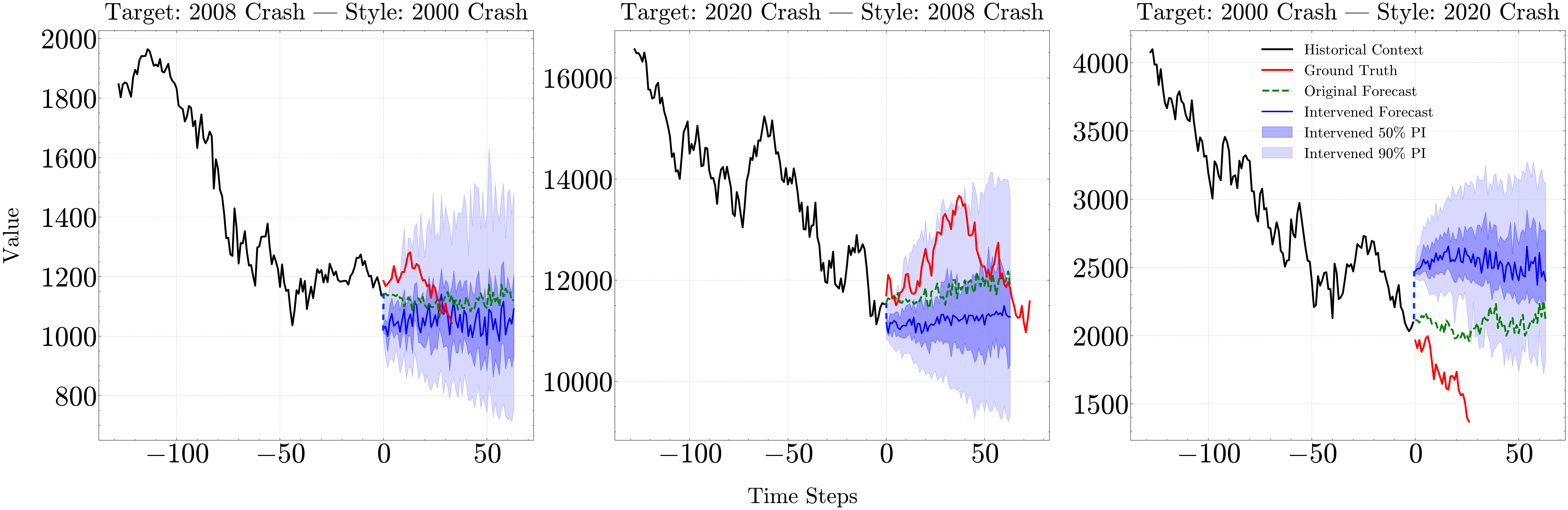}
        \caption{Cross-crash interventions reveal graded severity. Forecasts generated by transplanting crash signatures show that the forecast trajectory systematically deepens under severe signatures and is mitigated under milder ones, demonstrating that TSFMs encode crash events along a continuous latent severity axis.}
        \label{fig:toto_crash_crash}
    \end{minipage}
    \vspace{-0.5em}
\end{figure*}

\textbf{A unified \enquote{crash} concept emerges and solidifies across model depth}.
To further validate our findings, we project latent activation vectors from each layer onto the top 20 principal components via Principal Component Analysis (PCA) (Refer to Appendix \ref{sec:pca-ablation} for ablations), thereby isolating the dominant low-dimensional structure of the representations. Within this reduced subspace, we compute cosine similarities for both within-regime and cross-regime inputs. As shown in Figure~\ref{fig:heatmap-plots}, latent layers exhibit consistently high similarity in this principal subspace for similar regime, indicating that distinct market regimes are encoded in a shared, lower-dimensional core representation (Appendix \ref{heatmap-ablation} for ablations). This provides evidence that high-level event concepts, such as market crashes and calm periods, emerge as compact and geometrically coherent directions in the latent space. \textit{Because regime information concentrates in a low-dimensional core, transplanting activation statistics perturbs shared concept directions and steers forecasts to the implanted regime}.



\textbf{Learned concepts are nuanced and quantitatively encode event severity}. Having established that models distinguish calm and crash regimes in latent space, we next ask whether the notion of a “crash” is monolithic or admits fine-grained distinctions. To probe this, we transplant activation signatures across historical crashes (Figure~\ref{fig:toto_crash_crash}). These cross-crash interventions reveal that the Dot-com (2000) signature induces sharper declines than the 2008 crisis, while the milder 2020 signature mitigates downturns. This demonstrates that TSFMs organize market regimes within a continuous semantic space, where each crash corresponds to a distinct point with interpretable magnitude.

\section{Conclusion}
This study provides the first causal evidence that TSFMs encode abstract, steerable market regime concepts. Activation transplantation deterministically controls forecasts across architectures, with crash signatures inducing downturns and calm signatures restoring stability. Cross-crash interventions reveal a continuous severity axis, embedding historical events as addressable latent directions.Together, these findings move beyond post-hoc interpretation toward causal manipulation of hidden states, enabling steerable, risk-aware \enquote{what-if”} forecasting.

{
\small
\bibliographystyle{plainnat}
\bibliography{main_bib}
}

\newpage
\begin{appendices}

\section{Limitations and Future Work}
While our work provides a foundational proof-of-concept for semantic control, it represents the first step into an exciting research landscape. Researchers can investigate whether these steerable semantic subspaces are a universal property of sequence models, extending beyond Transformers to  architectures like State Space Models \cite{ssm1,ssm2} or LSTM based time series models \cite{lstm1, lstm2, lstm3}. Furthermore, our intervention technique opens the door to discovering a broader \enquote{semantic vocabulary} within TSFMs, probing for concepts like earnings surprises in finance or seizure onsets in EEG data. Our findings invite a new research program focused not just on what these models can predict, but on the rich, internal worlds they build to do so.

\section{Experimental Setup}

The experiments are designed to show the robustness and generalizability of our findings. This is reflected in our choice of models, data, and parameters.



\subsection{Models and Hardware}
To validate our claims across different architectural paradigms and scales, we employ two distinct foundation models. The primary model for our main figures is \textbf{Toto-Open-Base-1.0} \cite{toto}, a 103M parameter decoder-only Transformer. To confirm our findings are not an architectural artifact, we replicate key experiments on four variants of \textbf{Chronos-T5} \cite{chronos}, an encoder-decoder architecture, ranging from the 8M parameter \texttt{tiny} variant to the 710M parameter \texttt{large} model. All experiments were conducted on a single NVIDIA A6000 GPU to ensure a consistent and reproducible hardware environment. We performed all our intervtions

\subsection{Real-World Data}
For our primary experiments, we use daily closing values of the NASDAQ-100 index. We extract multiple historical periods to serve as both \texttt{Target} and \texttt{Style} inputs, detailed in Table \ref{tab:data_periods}. In Figure~\ref{fig:intervention_crash_calm}, the point \enquote{$0$} marks the end of the input and the start of the prediction. \textbf{The $0$ points for all subplots in Figure~\ref{fig:intervention_crash_calm} correspond to the \textbf{End Date} in Table \ref{tab:data_periods}.}

\begin{itemize}
    \item \textbf{Input Context Length:} To create a continuous time series suitable for the model and account for non-trading days (weekends, holidays), we fill frequency gaps by inserting missing dates and imputing their values using the previous day’s index. From this continuous, imputed series, we use a fixed input length of \textbf{128 time steps} (days). This choice is deliberate; the Toto model utilizes a patch size of 64, and an input length of 128 ensures the model processes exactly two full, non-overlapping patches, providing a clean and consistent representational structure for our interventions.
\end{itemize}

\begin{table}[h!]
\centering
\caption{Historical periods from the NASDAQ-100 index used for \texttt{Target} and \texttt{Style} inputs in our experiments.}
\label{tab:data_periods}
\begin{tabular}{lccc}
\toprule
\textbf{Regime Name} & \textbf{Semantic Type} & \textbf{Start Date} & \textbf{End Date} \\
\midrule
2017 Calm & Calm & 2017-01-12 & 2017-05-20 \\
2007 Calm & Calm & 2007-03-12 & 2007-07-18 \\
2019 Calm & Calm & 2019-06-01 & 2019-10-07 \\
\midrule
2008 Crash & Crash & 2008-07-25 & 2008-11-30 \\
2000 Crash & Crash & 2000-08-31 & 2001-01-06 \\
2020 Crash & Crash & 2020-01-30 & 2020-06-06 \\
\bottomrule
\end{tabular}
\end{table}

\textbf{Note:} The end dates in the above Table reflect the final day of the \textit{input context} fed to the model. For visualization purposes in our figures, we naturally use an extended segment of the time series, as can be seen in our code, to plot the subsequent ground truth against which our forecasts are evaluated.

\subsection {Controlled Synthetic Data}
To provide a controlled validation of our severity-encoding hypothesis (Section \ref{sec:exp}), we generate synthetic crash signals.

\begin{itemize}
    \item \textbf{Input Context Length:} For these experiments, we use a longer input context of \textbf{256 time steps}. This allows us to observe how the model's behavior adapts when presented with a longer historical context than in the primary experiments.
    \item \textbf{Sampling:} For each forecast, we generate \textbf{256 samples} from the model's output head. This ensures a rich, well-defined predictive distribution, allowing for a robust analysis of both the median forecast and its associated uncertainty.
\end{itemize}

\subsection*{Synthetic Crash and Calm Regimes Generation} 
To systematically probe model behavior under controlled conditions, we generate synthetic financial time series using a discrete-time jump--diffusion process, a discretized analogue of the Merton model~\citep{merton1976option,cont2004financial}. 

Let $S_t$ denote the price and $X_t=\log S_t$ the log-price. The dynamics are given by
\begin{equation}
X_{t+1} \;=\; X_t \;+\; \Big(\mu - \tfrac{1}{2}\sigma^2\Big) \;+\; \sigma \varepsilon_t \;+\; J_t, 
\qquad \varepsilon_t \sim \mathcal{N}(0,1),
\label{eq: synthetic time series}
\end{equation}
where the jump term is defined as
\begin{equation}
N_t \sim \mathrm{Poisson}(\lambda), 
\qquad 
J_t \,=\, \sum_{k=1}^{N_t} Z_{t,k}, 
\qquad 
Z_{t,k} \overset{\text{i.i.d.}}{\sim} \mathcal{N}(\mu_J,\sigma_J^2).
\end{equation}
Here $N_t$ is the random number of jumps that occur between $t$ and $t+1$. 
If $N_t=0$, then $J_t=0$ and the update is purely drift--diffusion; 
if $N_t=1$, one Gaussian jump $Z_{t,1}$ is added; 
if $N_t=2$, two independent shocks occur and $J_t = Z_{t,1}+Z_{t,2}$, and so on. 
Thus $N_t$ controls \emph{how many jumps occur}, while $Z_{t,k}$ controls \emph{how large each jump is}. 

\textbf{Calm regimes}: We define calm periods as stable markets with small positive drift, very low volatility, and no jumps:
\begin{equation}
\mu,\, \sigma,\, \lambda,\, \mu_J,\, \sigma_J 
= 2\times 10^{-4},\; 3\times 10^{-3},\; 0,\; 0,\; 0 .
\label{eq: parameter calm}
\end{equation}

\textbf{Crash regimes}: By contrast, crash periods are instantiated by scaling parameters with a severity factor $s$: increasing volatility, amplifying negative drift, and introducing rare but negative jumps. Formally,

\begin{equation}
\begin{aligned}
\mu(s) &= -8\times 10^{-4}\,s, & \sigma(s) &= 8\times 10^{-3}\,s, \\
\mu_J(s) &= -2\times 10^{-2}\,s, & \sigma_J(s) &= 10^{-2}\sqrt{s}, \\
\lambda(s) &= 5\times 10^{-2}\,s. &&
\end{aligned}
\label{eq: parameter crash}
\end{equation}

\begin{figure}[!ht]
    \centering
    \subfloat[Synthetic Calm Series]{\includegraphics[width=0.45\textwidth]{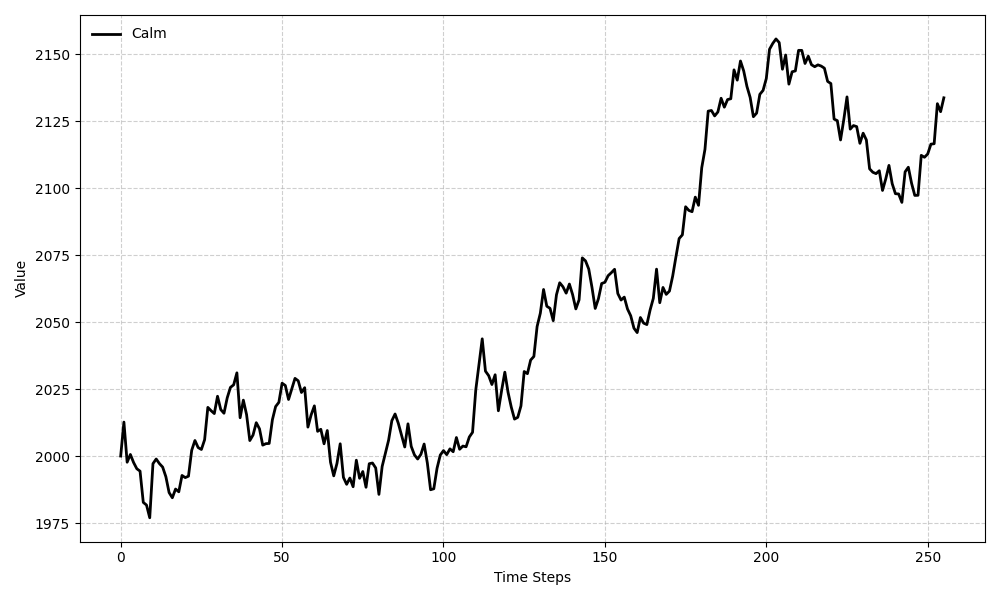}}
    \hfill
    \subfloat[Synthetic Crash Series]{\includegraphics[width=0.45\textwidth]{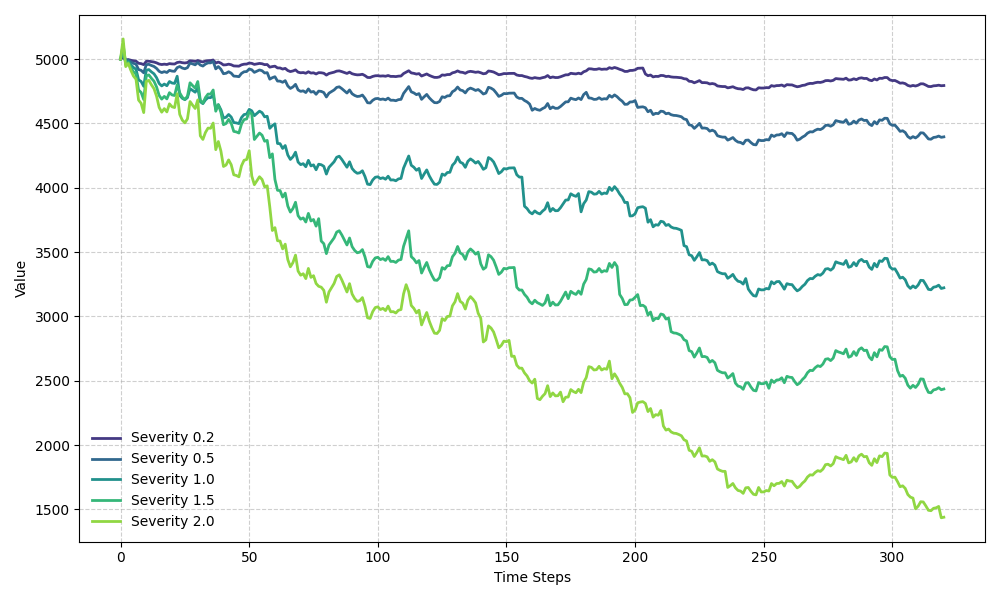}}
    \caption{\textbf{Synthetic series generation.} (a) Calm trajectory initialized at $X_0=2000$ and (b) crash trajectory initialized at $X_0=5000$, both generated using Eq.~\ref{eq: synthetic time series} with parameters specified in Eq.~\ref{eq: parameter calm} and Eq.~\ref{eq: parameter crash}, respectively.}

    \label{fig: synthetic_data}
    \vspace{-1em}
\end{figure}

This synthetic data framework provides a principled method for generating realistic calm and crash regimes, with a tunable severity parameter $s$ that controls the magnitude of systemic shocks. Figure~\ref{fig: synthetic_data} illustrates the synthetic calm and crash regimes used for the experiments in this work. By varying $s$, we obtain diverse univariate log-price time series that capture crashes of different intensities.

\section{Analysis on Synthetic data}

To further support our claims, we conduct detailed experiments using synthetic data, since real crash–calm data is not readily available. In this context, and to remain aligned with the financial domain, we generate synthetic datasets based on established techniques.

\subsection{Controlled Validation of Severity Encoding}
\label{sec:synth-ablation}
\textbf{Synthetic experiments confirm a continuous, severity-dependent encoding of crashes}. To isolate and rigorously test the model's nuanced understanding of crash severity observed in the main paper, we designed a controlled experiment using synthetic data. This allows us to move beyond the discrete, historical examples of the 2000, 2008, and 2020 crashes and probe the model's response to a continuous spectrum of event magnitudes.

\begin{figure}[!ht]
    \centering
    \includegraphics[width=0.7\textwidth]{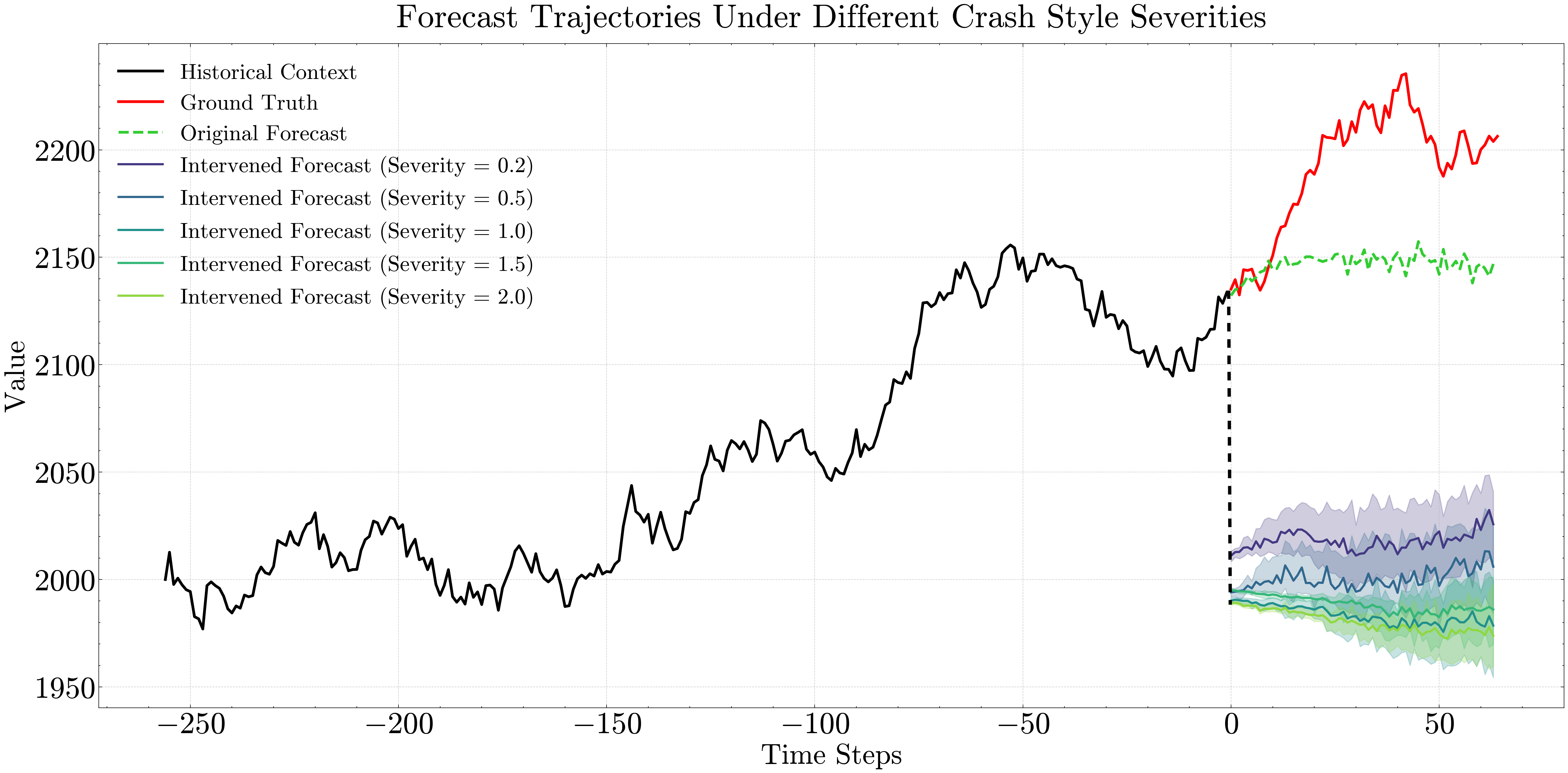}
    \caption{\textbf{Synthetic severity interventions for Toto-Open-Base-1.0}. Forecasts generated by transplanting activation statistics from synthetic crash signals of increasing intensity (Severity = 0.2–2.0) into a calm context. The forecasts exhibit a clear severity–response relationship, with downturn magnitude and predictive uncertainty both scaling with severity.}
    \label{fig:severity_spectrum}
    \vspace{-1em}
\end{figure}

Figure~\ref{fig:severity_spectrum} shows the result of intervening on a calm historical context by transplanting activation statistics from synthetically generated crash signals of varying intensity, from a mild Severity = 0.2 to an extreme Severity = 2.0. The intervened forecasts exhibit a perfect dose-response relationship: a low severity of 0.2 induces only a mild, stabilizing downturn relative to the baseline, whereas increasing the severity to 1.0 and 2.0 produces progressively steeper and more dramatic forecasted declines. The clear visual ordering of the forecasts further highlights the model's ability to interpret this quantitative information.

This experiment provides two crucial insights. 
\begin{enumerate}[leftmargin=1cm]
    \item Model's internal representation of a \enquote{crash} is not a binary switch but a continuous, quantifiable concept. The intervention's magnitude directly and proportionally controls the forecasted outcome, confirming that our hypothesis of the model learning from a continuous space of severity is correct.
    \item the prediction intervals also widen with increasing severity. This mirrors our finding with real-world data and reveals a sophisticated and consistent model behavior: more extreme interventions, which push the internal state further from its original context, correctly result in higher aleatoric uncertainty about the future path.
\end{enumerate}

\subsection{Synthetic Validation of Cross-Regime Representational Geometry}
\label{heatmap-ablation}

\textbf{Cosine Similarity:} Cosine similarity measures the cosine of the angle between two vectors, providing a simple yet effective way to assess similarity. Given two layer activation matrices $A^{(l)}$ and $A^{(m)}$, representing activations from layers $l$ and $m$, the cosine similarity is computed as:  

\begin{equation}
\text{cosine\_similarity}(\mathbf{A}^{(l)}, \mathbf{A}^{(m)}) = 
\frac{1}{n} \sum_{i=1}^{n} 
\frac{\mathbf{a}^{(l)}_i \cdot \mathbf{a}^{(m)}_i}{\|\mathbf{a}^{(l)}_i\|\|\mathbf{a}^{(m)}_i\|}
\label{eq:cosine}
\end{equation}

where $\mathbf{a}^{(l)}_i$ and $\mathbf{a}^{(m)}_i$ are the $i$-th activation vectors from layers $l$ and $m$, and $n$ is the number of samples.

We extend the controlled experiment by pairing the previously generated synthetic crash sequences with corresponding synthetic calm data. The goal is to test whether the representational geometry observed with real events generalizes beyond specific historical contexts.  

\begin{figure}[!ht]
    \centering
    \subfloat[Calm $\times$ ($s=0.2$)]{
        \includegraphics[width=0.22\textwidth]{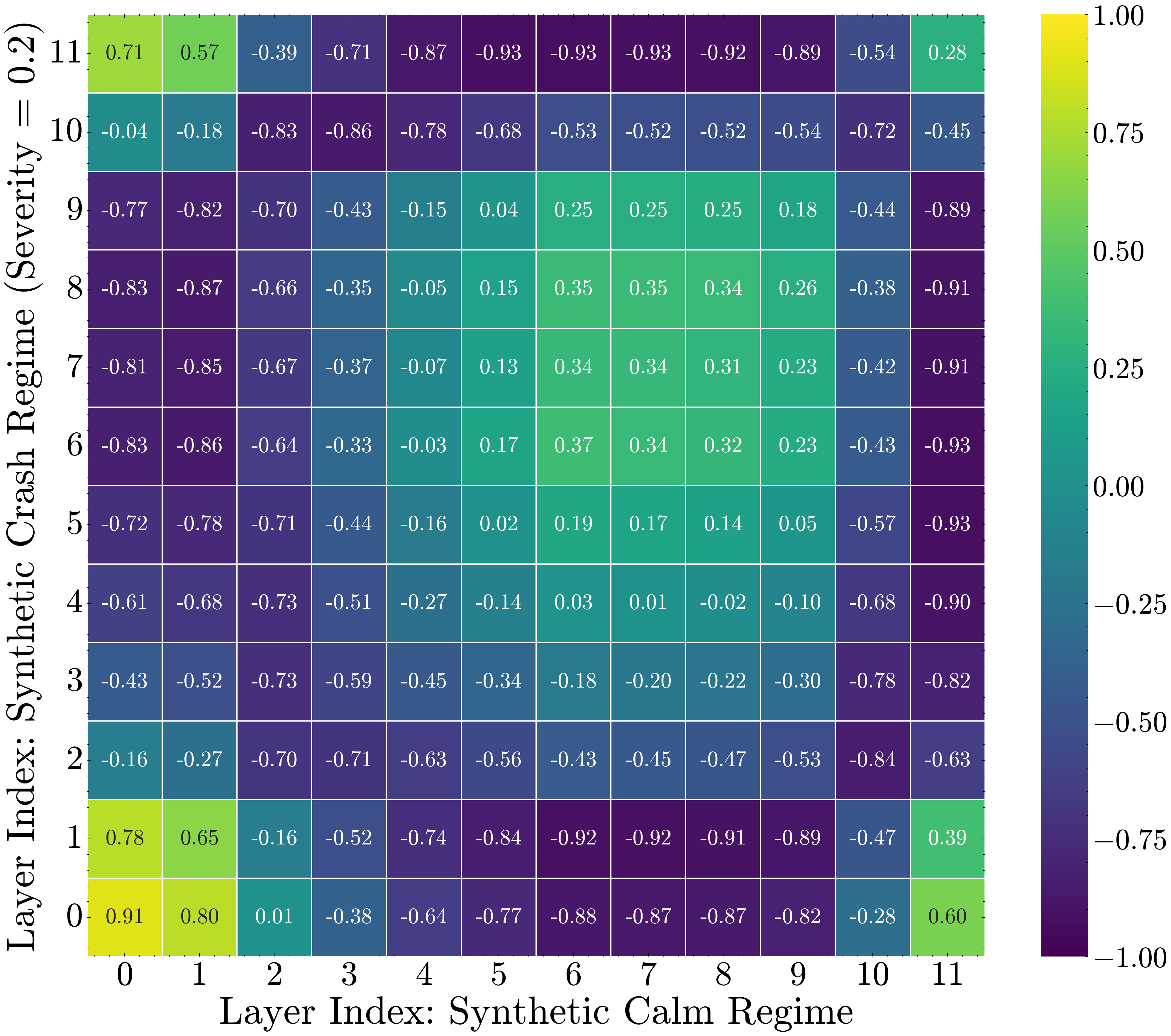}
    }
    \hfill
    \subfloat[Calm $\times$ ($s=1.0$)]{
        \includegraphics[width=0.22\textwidth]{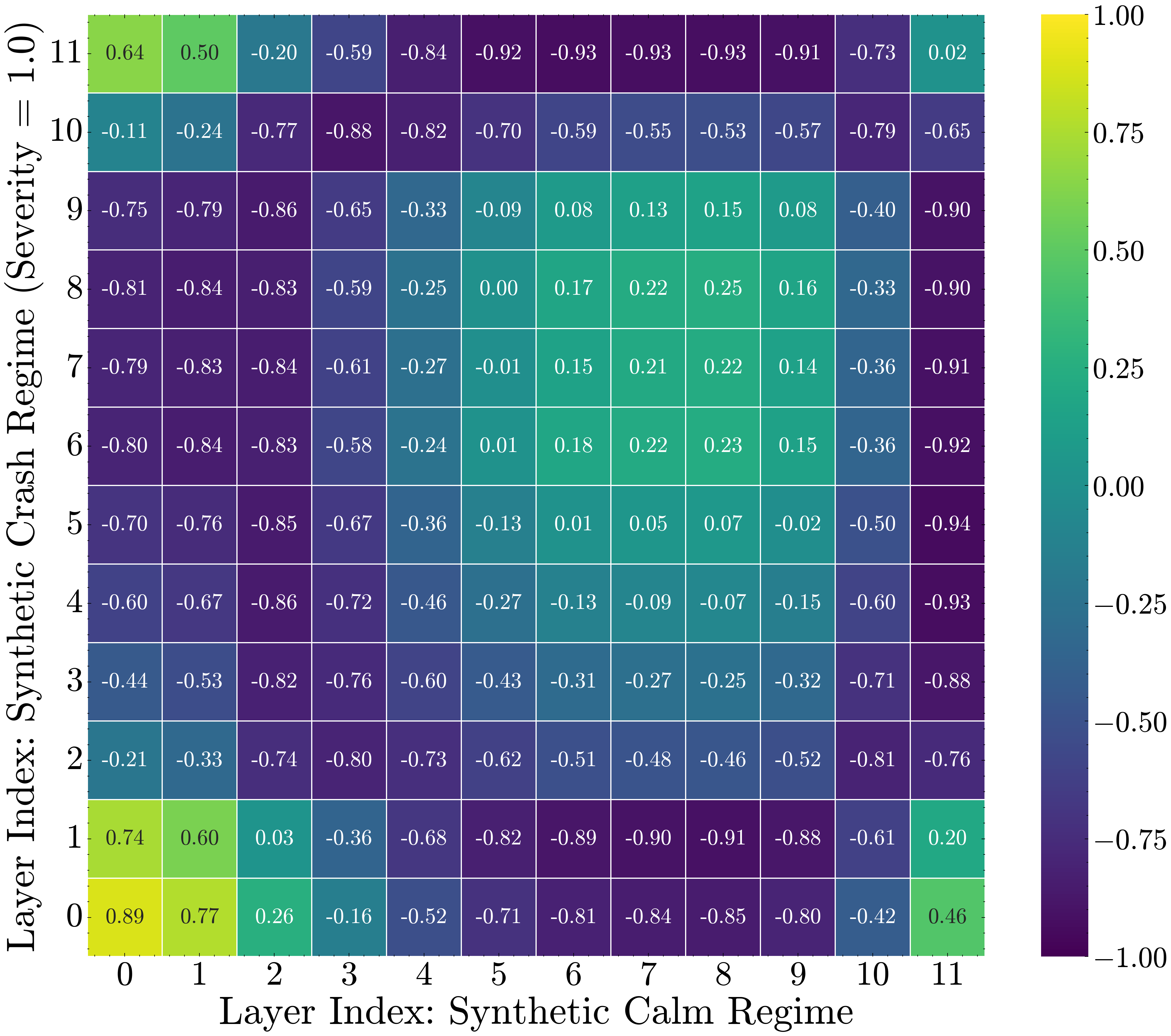}
    }
    \hfill
    \subfloat[Calm $\times$ ($s=2.0$)]{
        \includegraphics[width=0.22\textwidth]{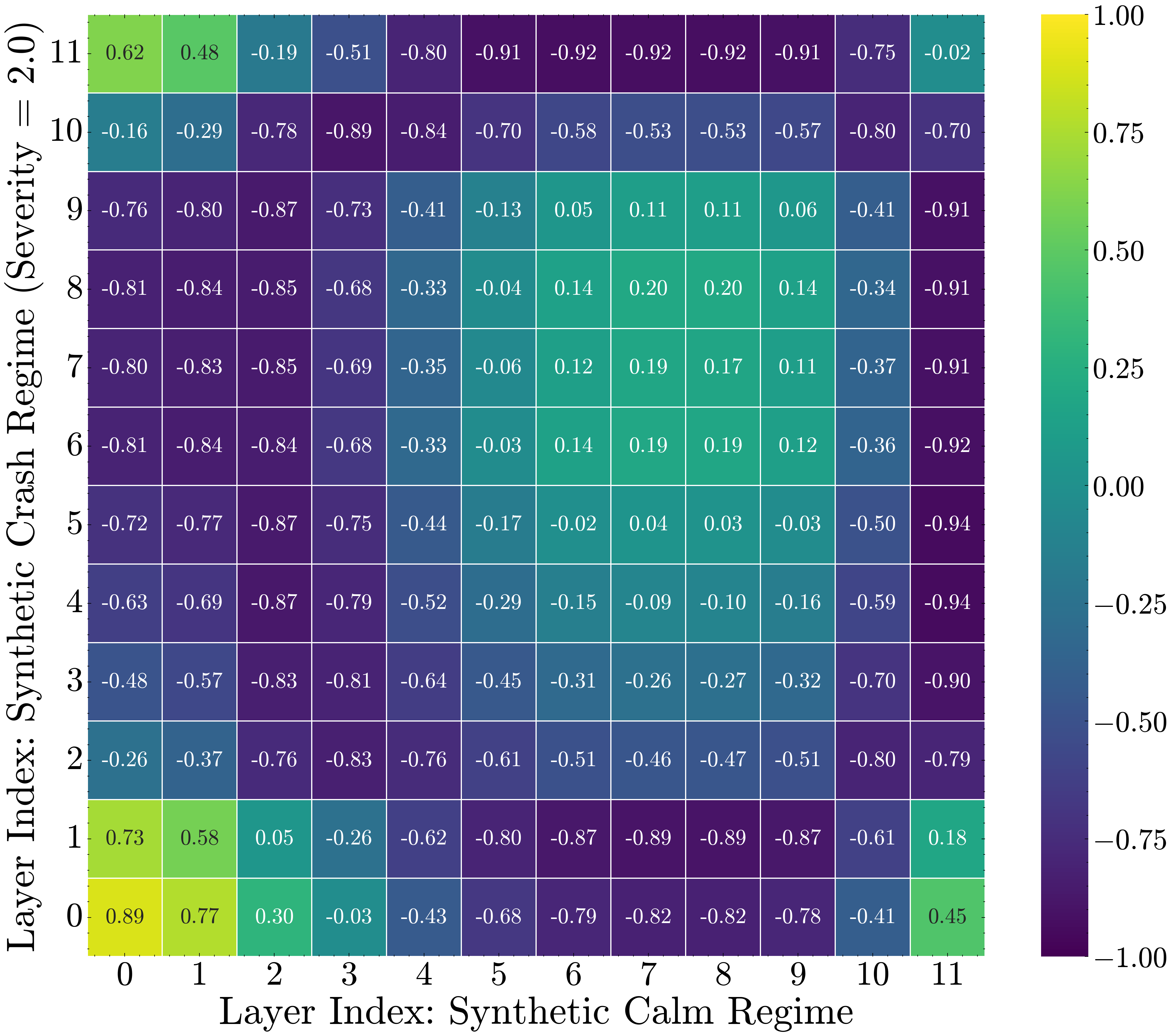}
    }
    \\
    \subfloat[($s=0.2)\times (s=1.0$)]{
        \includegraphics[width=0.22\textwidth]{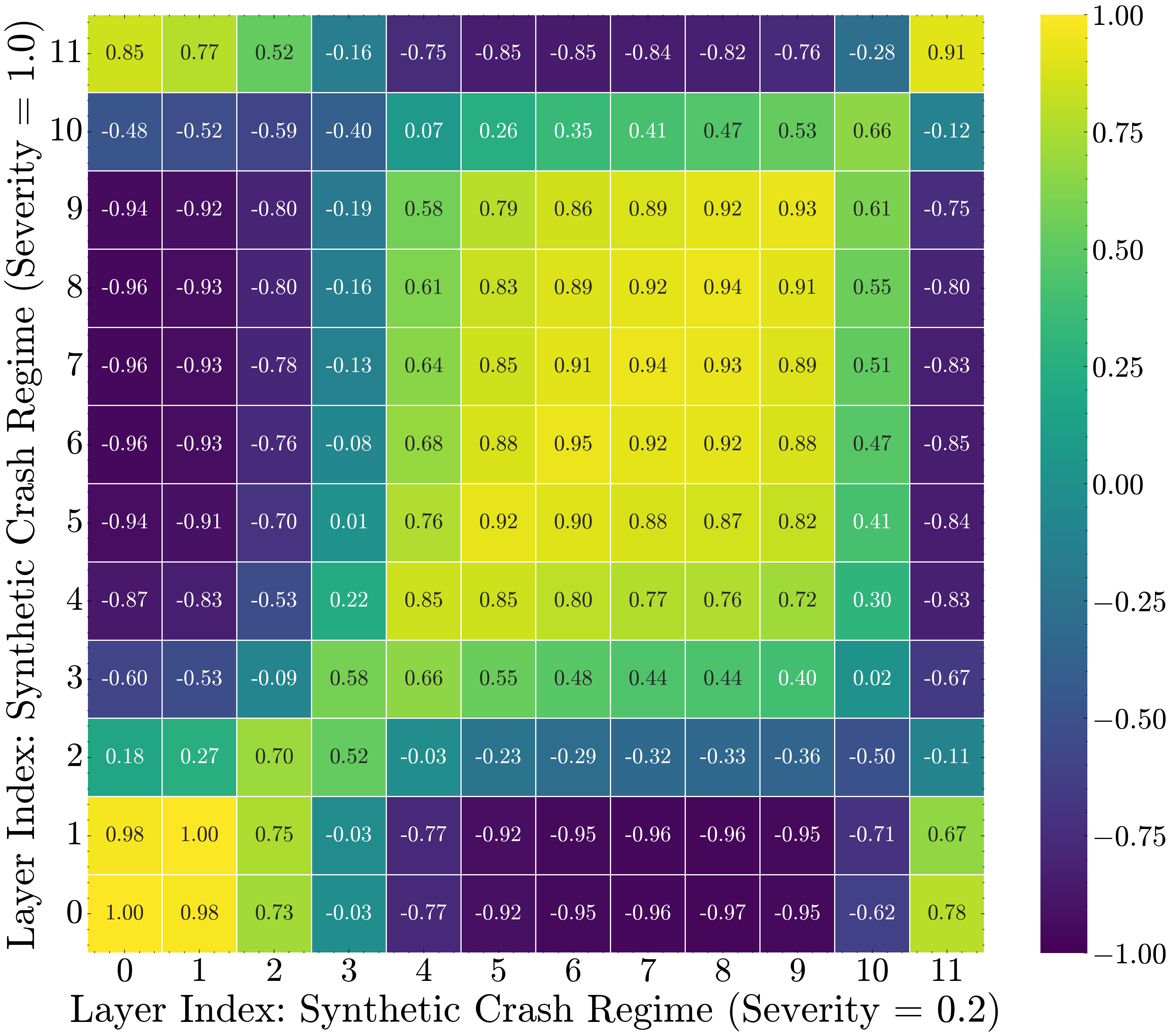}}
    \hfill
    \subfloat[$(s=0.2) \times (s=1.5)$]{
        \includegraphics[width=0.22\textwidth]{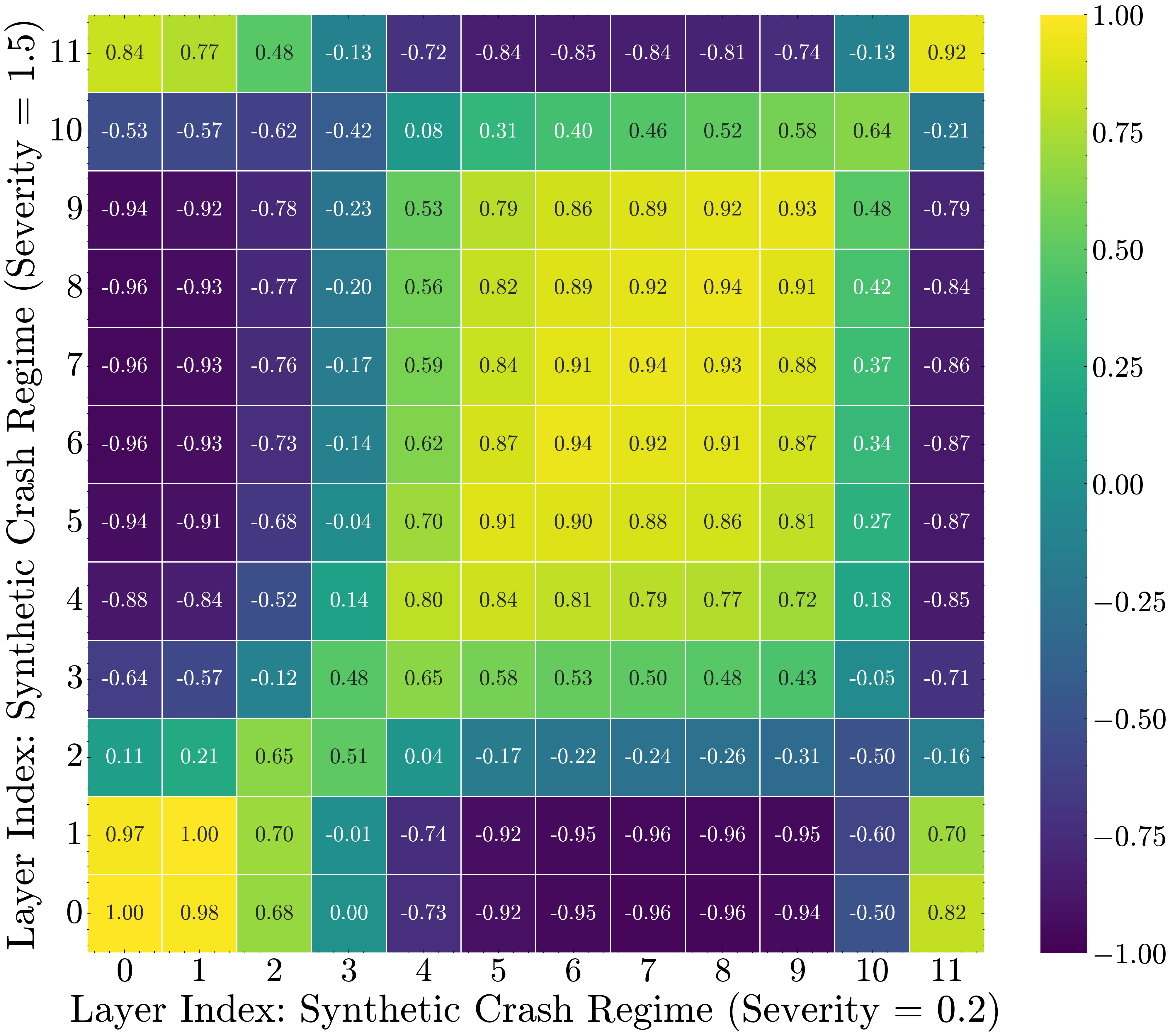}
    }
    \hfill
    \subfloat[$(s=0.2) \times (s=2.0)$]{
        \includegraphics[width=0.22\textwidth]{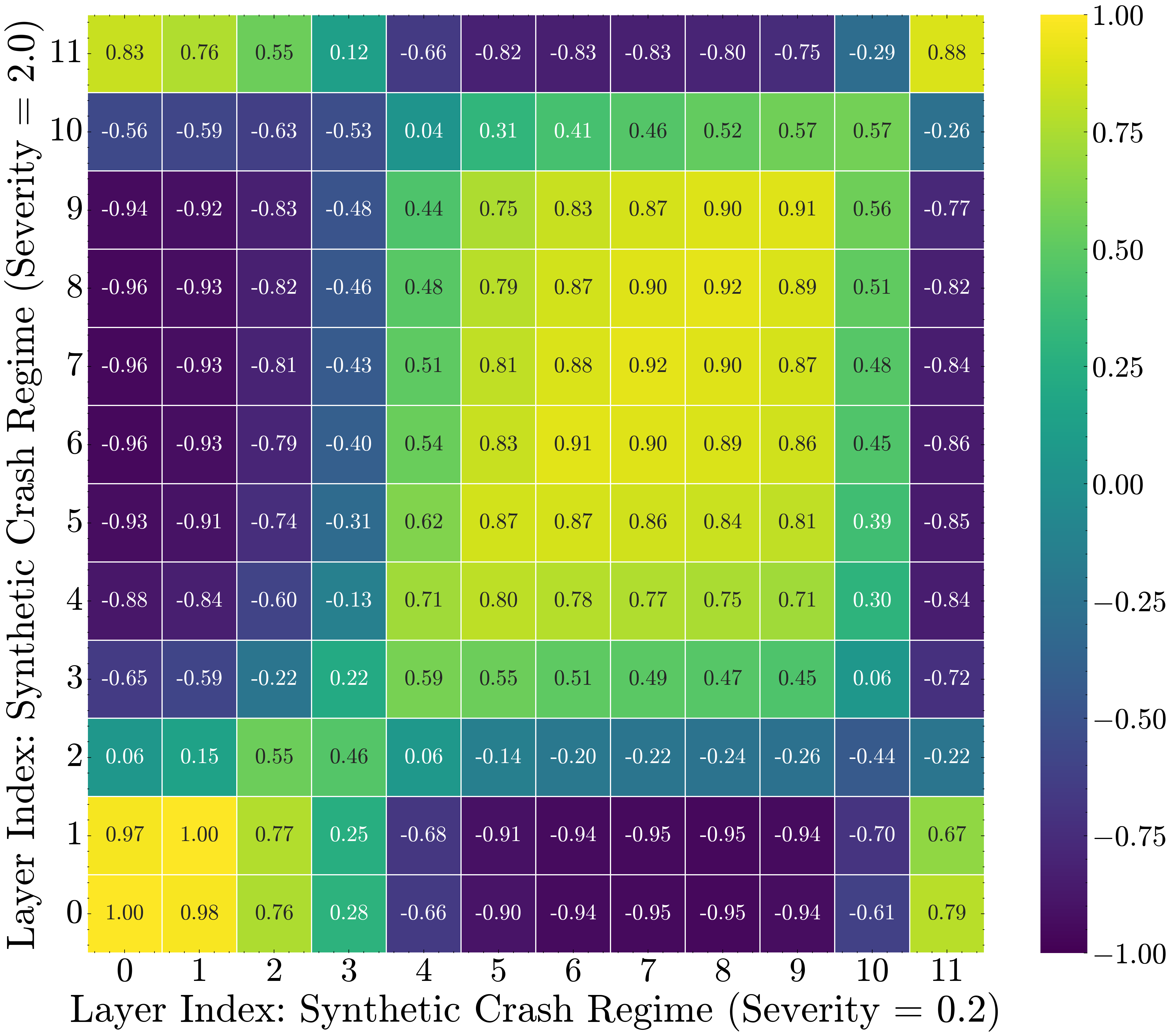}
    }
    \caption{Synthetic regime similarity. Cosine similarity matrices in the PCA-reduced latent subspace (20 components) of Toto-Open-Base-1.0. Top row: cross-regime comparisons between synthetic crash and calm sequences at increasing severity levels ($s=0.2,1.0,2.0$). Bottom row: within-regime comparisons across synthetic crash sequences of different severity.}
    \label{fig:synth-heatmaps}
    \vspace{-1em}
\end{figure}

\textbf{Cross- and inter-regime representational similarity.}  
Figure~\ref{fig:synth-heatmaps} shows cosine similarity matrices between synthetic crash regimes of varying severity and a calm baseline, projected into the PCA-reduced latent subspace of the Toto-Open-Base-1.0 model. As severity ($s$) increases, cross-regime similarity decreases consistently across depth, indicating that the model separates mild from extreme crashes more distinctly in latent space. This mirrors our real-world findings: just as historical crashes align along a continuous severity axis, synthetic crashes exhibit progressively lower similarity to calm regimes as intensity grows. These results confirm that severity encoding is not tied to idiosyncratic market episodes, but reflects a generalizable and quantifiable mechanism for representing systemic shocks.

\textbf{Within-regime representational similarity.}  
We next compare synthetic crash regimes of increasing severity against a mild-severity baseline ($s=0.2$). Although similarity gradually decreases as the severity gap widens, the correlations remain consistently high across layers. This indicates that while the model is sensitive to graded differences in severity, all crash regimes are still anchored within a coherent latent subspace. In other words, the notion of a ``crash'' is stable and abstract, with severity expressed as a continuous modulation rather than a categorical separation.


\section{Causal Interventions are Invariant of Model-Size}\label{sec: model invariance}
\label{sec:chronos-ablations}
\begin{figure}[!ht]
    \centering
    \subfloat[tiny (8M)]{\includegraphics[width=0.23\textwidth]{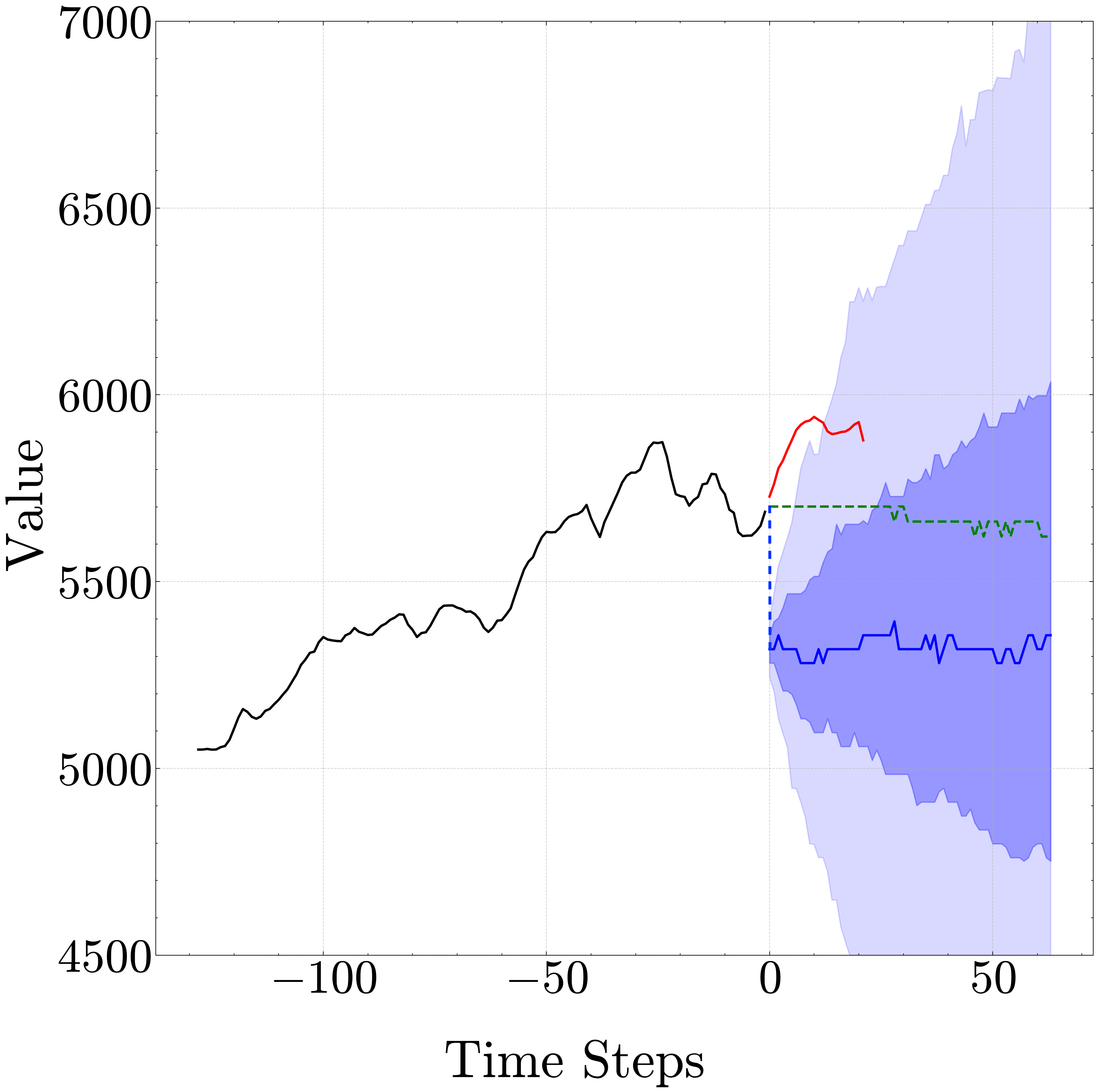}
    }
    \hfill
    \subfloat[small (46M)]{\includegraphics[width=0.23\textwidth]{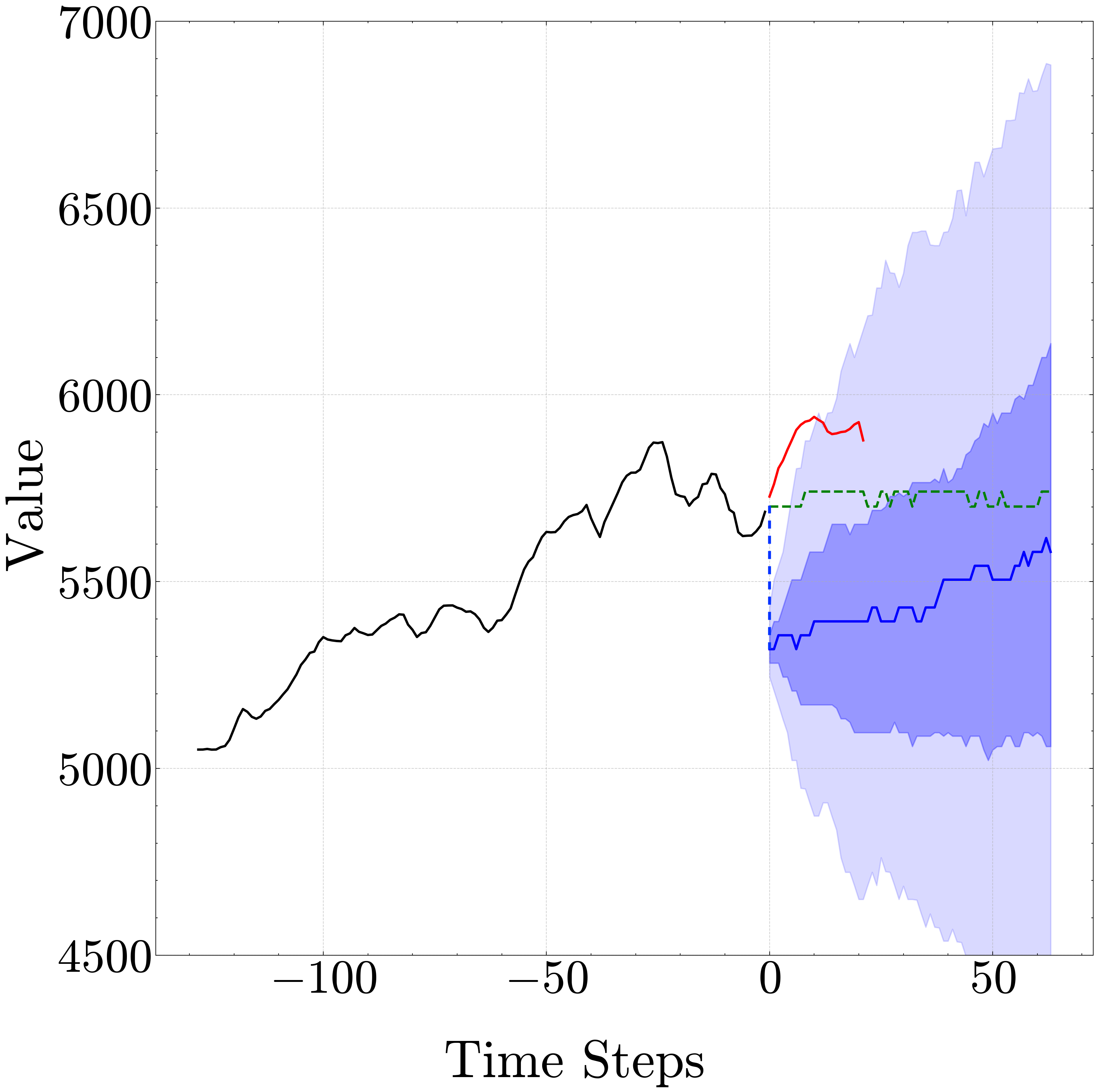}
    }
    \hfill
    \subfloat[base (200M)]{\includegraphics[width=0.23\textwidth]{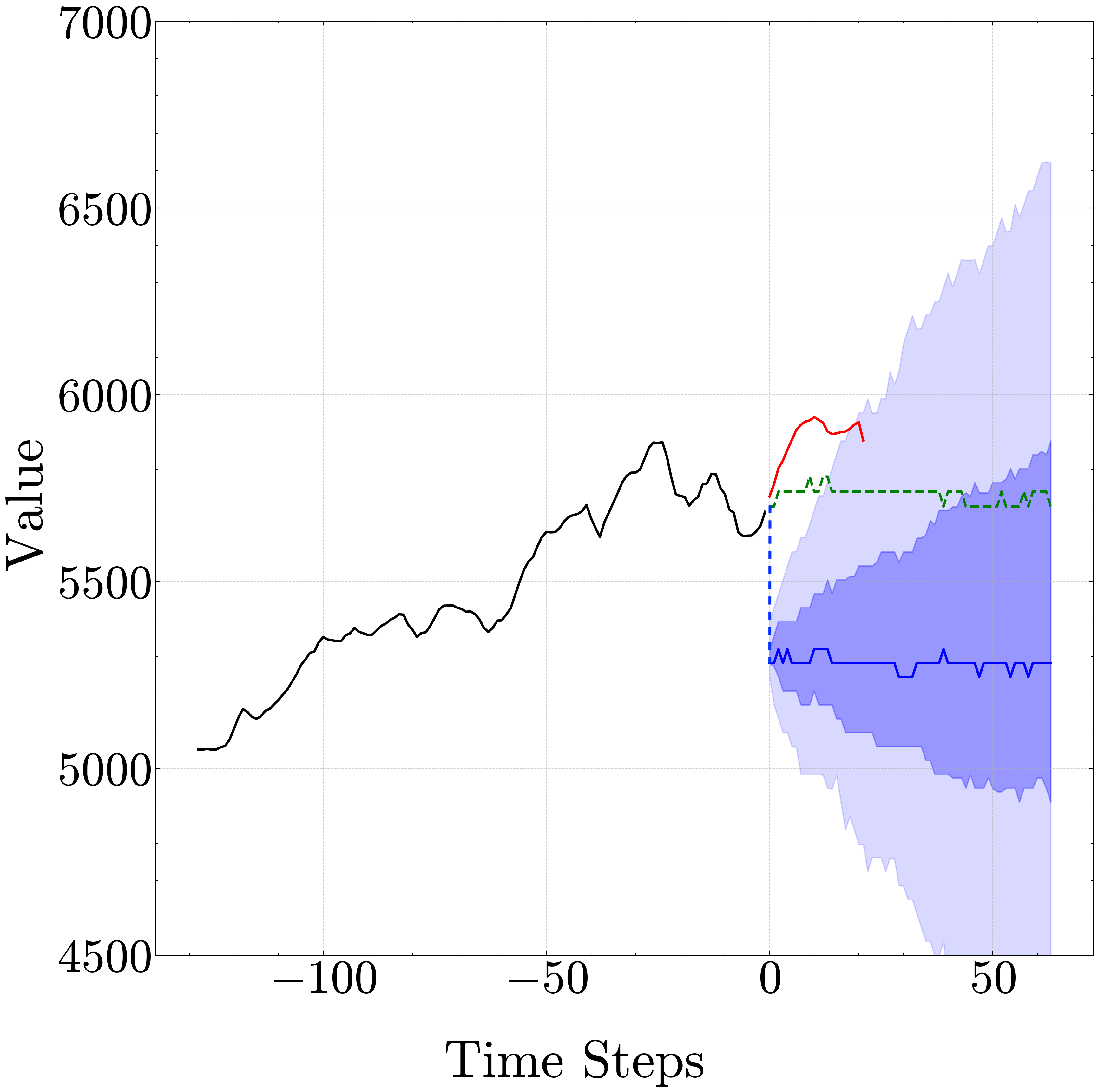}
    }
    \hfill
    \subfloat[large (710M)]{        \includegraphics[width=0.23\textwidth]{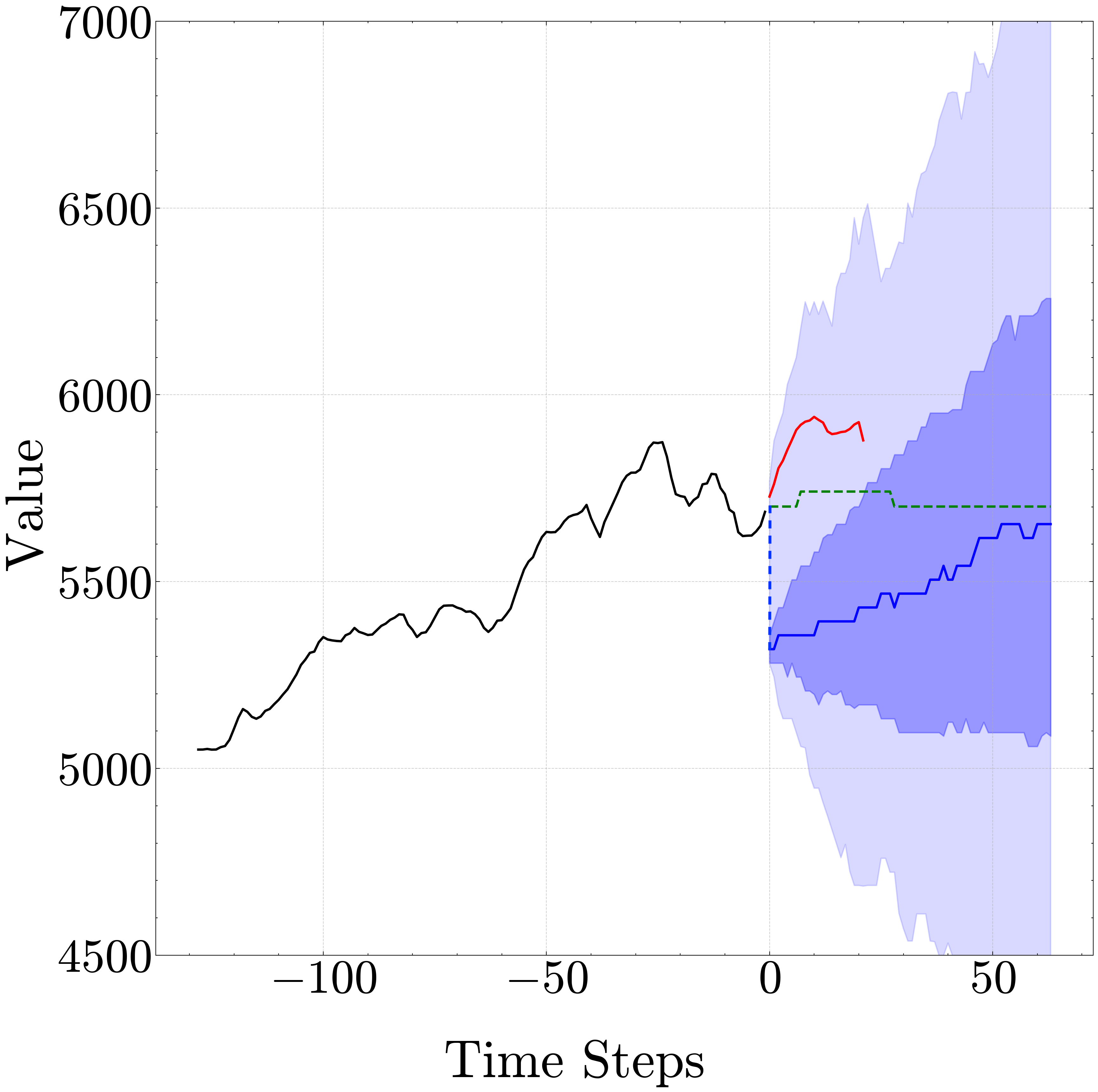}
    }
    \caption{\textbf{Chronos model interventions across scales:} Intervened forecasts for tiny, small, base, and large Chronos variants with 2008 crash semantics transplanted into 2017 calm data. The number of model parameters is indicated in parentheses for each variant. }
    \label{fig:chronos-size-ablation}
    \vspace{-1em}
\end{figure}

Since the Toto-Open-Base-1.0 model is released only in a single configuration, we perform model size ablations using Chronos, which is available in multiple variants. As shown in Figure~\ref{fig:chronos-size-ablation}, activation transplantation yields consistent interventions across all Chronos variants, demonstrating that causal controllability is invariant to model size and thus reflects a fundamental property of time series foundation models, rather than a capacity artifact.

\section{Activation-Space Similarity Across Model Depth}

\textbf{A unified \enquote{crash} concept emerges and solidifies across model depth}. In addition to analyzing representations in the PCA-reduced subspace, \textbf{we also evaluate similarity directly in the original activation layers}. To probe the mechanism that enables our interventions, we quantify the representational geometry of market regimes using cosine similarity across layers (Table~\ref{tab:cosine_similarity}). Early layers show strong negative correlation between crash and calm regimes, capturing their antagonistic nature at the level of local features. By mid-depth (e.g., Layer 5), distinct crash events begin to collapse into a shared subspace, and in the final layers they converge into a highly aligned representation. This progression indicates the emergence of a stable and abstract “crash” concept, which provides a consistent semantic anchor for causal interventions to operate upon.

\begin{table}[!ht]  
\centering
\footnotesize
\caption{Representational similarity: Cosine similarities between latent activation vectors across layers for cross-regime inputs for Toto model.}
\label{tab:cosine_similarity}
\begin{tabular}{l|ccccc}
\toprule
\textbf{Events} & \textbf{L1} & \textbf{L2} & \textbf{L5} & \textbf{L9} & \textbf{L11} \\
\midrule
$2008$ (Crash) $\times$ $2017$ (Calm) & -0.311 & -0.194 & 0.327 & 0.562 & 0.741 \\
$2000$ (Crash) $\times$ $2019$ (Calm) & -0.416 & -0.228 & 0.240 & 0.533 & 0.684 \\
$2000$ (Crash) $\times$ $2008$ (Crash) & 0.193 & 0.471 & \textbf{0.915} & \textbf{0.932} & \textbf{0.968} \\
$2008$ (Crash) $\times$ $2020$ (Crash) & 0.152 & 0.439 & \textbf{0.882} & \textbf{0.938} & \textbf{0.957} \\
\bottomrule
\end{tabular}
\vspace{-2mm} 
\end{table}

\section{Ablation on Dimensionality Reduction for Similarity Analysis
}
\label{sec:pca-ablation}
\begin{figure}[!ht]
    \centering
    \subfloat[$k=20$]{\includegraphics[width=0.21\textwidth]{fig/calm_vs_crash_heatmap.png}}
    \hfill
    \subfloat[$k=30$]{\includegraphics[width=0.21\textwidth]{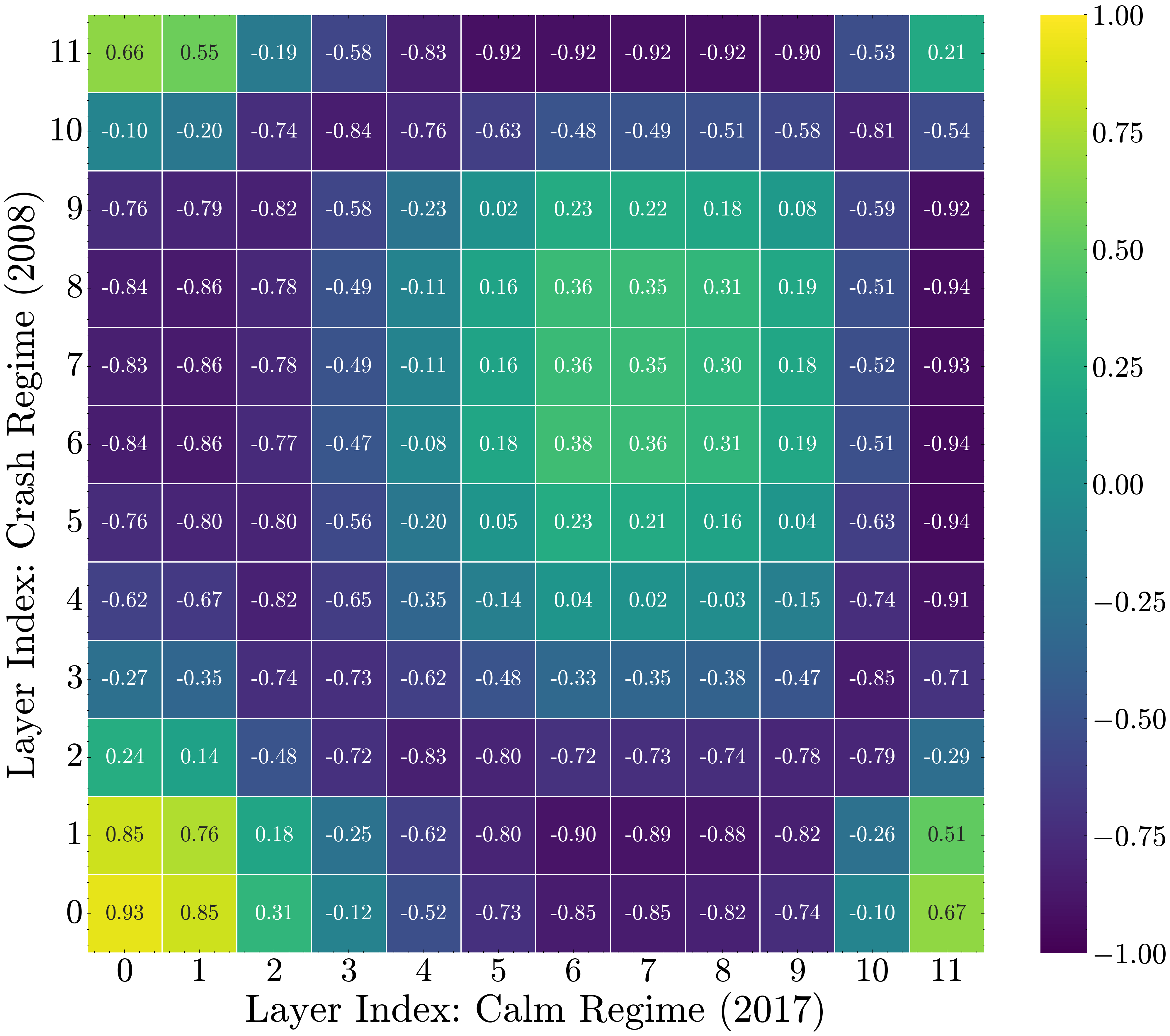}}
    \hfill
    \subfloat[$k=40$]{\includegraphics[width=0.21\textwidth]{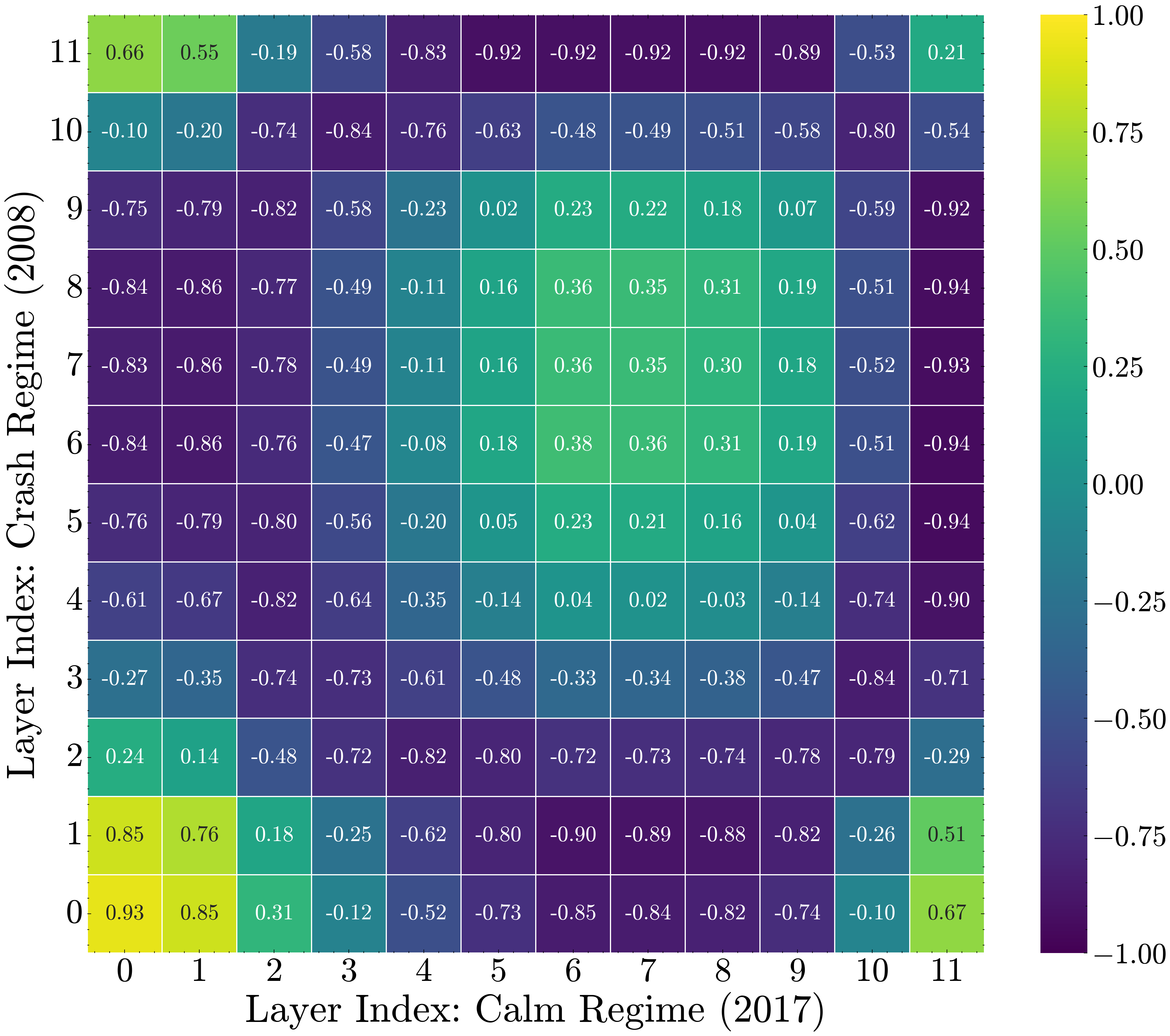}}
    \hfill
    \subfloat[$k=50$]{\includegraphics[width=0.21\textwidth]{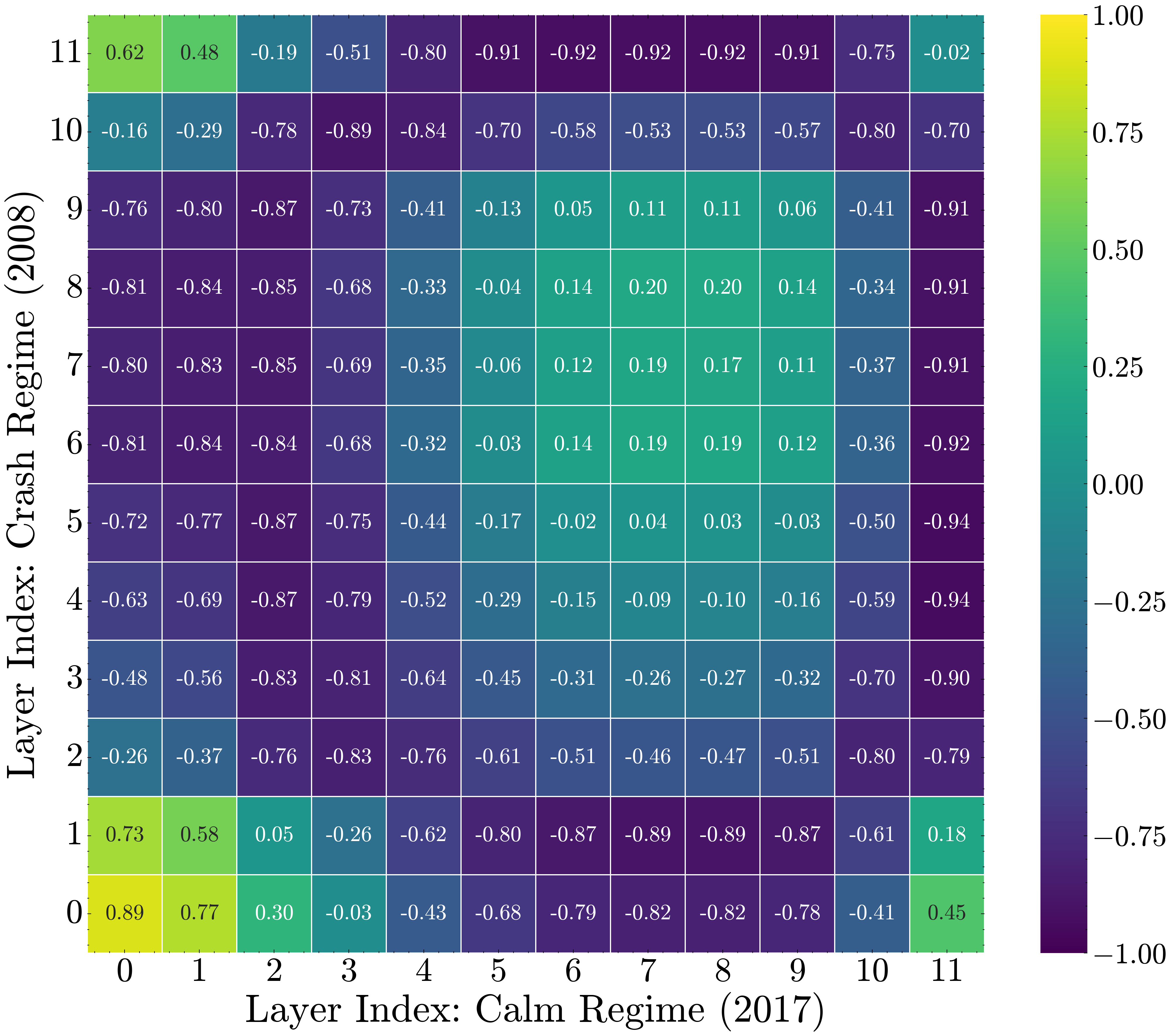}}
    \caption{\textbf{PCA ablation on representational similarity.} Cosine similarity matrices for crash–calm regimes under different choices of principal components. Increasing $k$ beyond 20 progressively diffuses the structured separation between regimes, supporting our choice of $k=20$ in the main analysis.}
    \label{fig: PCA ablation}
    \vspace{-1em}
\end{figure}

\textbf{The choice of PCA components is critical for isolating the robust semantic signal from high-frequency noise}. A core component of our analysis is measuring similarity between high-dimensional activation vectors. To do this robustly, we first project the activations onto their top $k$ principal components. The choice of k is a crucial hyperparameter: too low, and we risk losing the core semantic information; too high, and we risk including instance-specific noise that obscures the underlying conceptual structure.
Figure~\ref{fig: PCA ablation} provides the justification for our choice of $k$=20 in the main paper by showing the results for higher values: $k$=30, $k$=40, and k=50. The analysis reveals a clear and consistent trend: as we include more principal components beyond our chosen threshold, the clean, interpretable structure of the similarity matrix begins to diffuse.

We observed in Figures \ref{fig:heatmap-plots} and \ref{fig:synth-heatmaps} that early layers show anti-correlation, mid-layers transition, and late layers achieve stable orthogonality. This clear pattern becomes progressively corrupted with noisy artifacts. This is not because the higher components are meaningless, but because they capture finer-grained, time-step-specific variance that is irrelevant to the high-level, abstract concept of \enquote{crash} or \enquote{calm}.

This ablation demonstrates that the core semantic difference between market regimes is a relatively low-rank signal, effectively captured within the top 20 principal components. Our choice of $k$=20 is therefore a principled one, designed to maximize signal fidelity by capturing the entirety of the shared semantic concept while filtering out the high-frequency noise that varies from one specific market day to the next. This ensures that our main results reflect the model's understanding of the abstract concept itself, not the quirks of a particular historical instance.

\end{appendices}

\end{document}